%% file: iclr2027_conference.tex
\documentclass{article} 
\usepackage[T1]{fontenc}
\usepackage{iclr2027_conference,times}

\input{math_commands.tex}

\usepackage{amssymb}

\usepackage{hyperref}
\usepackage{url}
\usepackage{graphicx}
\usepackage{xcolor}
\usepackage{colortbl}
\usepackage{wrapfig}
\usepackage{needspace}
\usepackage{booktabs,multirow}
\usepackage{listings}
\usepackage[most]{tcolorbox}

\definecolor{PromptFrame}{HTML}{C3CBD6}
\definecolor{PromptBack}{HTML}{F7F9FB}
\definecolor{PromptSlot}{HTML}{1F5FA8}
\lstdefinestyle{promptstyle}{
  basicstyle=\ttfamily\fontsize{7.6}{9.1}\selectfont,
  breaklines=true,
  breakindent=0pt,
  breakautoindent=false,
  postbreak=\mbox{\textcolor{PromptFrame}{$\hookrightarrow$}\space},
  columns=fullflexible,
  keepspaces=true,
  upquote=true,
  showstringspaces=false,
  aboveskip=0pt,
  belowskip=0pt,
  moredelim=**[s][\color{PromptSlot}]{\{}{\}},
}
\newtcblisting{promptbox}[1]{
  breakable, enhanced,
  listing only, listing style=promptstyle,
  colback=PromptBack, colframe=PromptFrame,
  boxrule=0.4pt, arc=1.6pt,
  left=4pt, right=3pt, top=3.5pt, bottom=3.5pt,
  toptitle=1pt, bottomtitle=1pt,
  fonttitle=\sffamily\bfseries\fontsize{7.4}{8.6}\selectfont,
  coltitle=black, colbacktitle=PromptBack,
  title={#1},
  attach boxed title to top left={xshift=5pt, yshift=-\tcboxedtitleheight/2},
  boxed title style={colback=white, colframe=PromptFrame, boxrule=0.4pt,
                     arc=1pt, left=2.5pt, right=2.5pt, top=0.6pt, bottom=0.6pt},
}

\title{Draft-KV: Learning Useful Latent Communication Between Language Models}

\author{{\bfseries Linquan Wu$^{1}$, Shichang Meng$^{1}$, Tianxiang Jiang$^{2}$,
Haoyu Yang$^{3}$, Peng Zhong$^{4}$,} \\
{\bfseries Fengming Zhu$^{4}$, Xi Peng$^{5}$, Linqi Song$^{1}$, Jacky Keung$^{1}$,
Jingyu Zhang$^{6}$} \\[0.8em]
$^{1}$City University of Hong Kong \\
$^{2}$University of Science and Technology of China \\
$^{3}$University of Electronic Science and Technology of China \\
$^{4}$AIPD, Tencent \\
$^{5}$Theory Lab, Huawei \\
$^{6}$Hong Kong Metropolitan University \\[0.7em]
\href{https://github.com/Svardfox/Draft-KV}{\raisebox{-0.18em}{\includegraphics[height=1.05em]{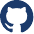}}\hspace{0.4em}{\mdseries\ttfamily\color[HTML]{1B3A6B}github.com/Svardfox/Draft-KV}}
}

\newcommand{\TFLOPS}{148}

\definecolor{accentink}{HTML}{1B3A6B}

\newenvironment{takeaway}
  {\par\addvspace{5pt}%
   \noindent\textcolor{accentink}{\rule{\linewidth}{0.8pt}}\par\nobreak\vspace{1pt}%
   \noindent\textcolor{accentink}{\scshape Takeaway.}\enspace\ignorespaces}
  {\par\nobreak\vspace{1pt}%
   \noindent\textcolor{accentink}{\rule{\linewidth}{0.3pt}}\par
   \addvspace{5pt}}

\iclrfinalcopy
\begin{document}
\raggedbottom

\maketitle
\lhead{}

\begin{abstract}
Latent communication passes internal states between language models instead of
decoded text, but higher receiver accuracy does not show that the receiver used
the message content. Across five method--dataset pairs, replacing each message
with one from an unrelated question changes accuracy by at most $0.60$ points,
even when communication adds $15.44$ points over the receiver alone. Thus the
interface can supply the gain while making the sharer dispensable.
\textbf{Draft-KV} instead sends the key--value states formed while the sharer
drafts an answer to the current question. Linear projections place these states
in a side memory read through a gated attention branch, and progressive training
moves from message reconstruction to answer supervision under a guard on harm
from mismatched messages. Both models remain frozen and the interface trains
$1.05$M parameters, $348\times$ fewer than C2C. With a Qwen3-8B sharer, a frozen
Qwen2.5-0.5B-Instruct receiver reaches $78.04\%$ on MMLU-Redux, versus $37.45\%$
alone and $36.40\%$ with reassigned messages. At fixed interface size, scaling
the sharer from $0.6$B to $8$B raises accuracy from $46.11\%$ to $78.04\%$;
communication also transfers to held-out tasks and can exceed both models when
each holds different evidence.
\end{abstract}

\section{Introduction}
\label{sec:introduction}

\input{sections/introduction}

\section{Related Work}
\label{sec:related-work}

\input{sections/related_work}

\section{The Information-Use Gap in Latent Communication}
\label{sec:information-use-gap}

\input{sections/information_use_gap}

\section{Draft-KV}
\label{sec:method}

\input{sections/method}

\section{Experiments}
\label{sec:experiments}

\subsection{Experimental Setup}
\label{sec:experimental-setup}

\input{sections/experiments_setup}

\subsection{Main Results}
\input{sections/main_results_table}
\input{sections/main_results}

\subsection{Message Intervention Analysis}
\label{sec:message-intervention}
\input{sections/message_intervention}

\subsection{Sharer Scaling}
\label{sec:sharer-scaling}
\input{sections/sharer_scaling}

\subsection{Ablation Study}
\label{sec:ablation}
\input{sections/ablation_study}

\section{Conclusion}
\label{sec:conclusion}

Latent communication can raise accuracy without carrying content: in the interfaces we audit,
swapping in another question's message barely changes accuracy. Draft-KV instead sends the
key--value states formed while the sharer drafts and trains the receiver to use them. Its gains grow
with sharer capability and disappear under message reassignment, making pairing gain---not accuracy
alone---the evidence that communication carries useful content.

\subsection*{AI Use Statement}

We used a general-purpose large language model assistant in two roles, both
under author review. For writing, it helped with copy-editing, LaTeX
formatting, and tightening prose in the main text and the appendices. For code,
it helped draft parts of the evaluation and plotting scripts, which we then
reviewed and re-ran end to end. We did not use generative AI to propose the
research question, to design the message-intervention protocol, or to produce
any reported number; all accuracies, gains, and diagnostics come from our own
runs on the checkpoints and evaluation splits documented in
Appendix~\ref{app:experimental-setup}. We take responsibility for the final
content of this work.

\subsection*{Ethics Statement}

This study raises no ethical concerns: all experiments use publicly released
models and public benchmarks, and involve no human subjects and no sensitive
personal data.

\subsection*{Reproducibility Statement}

Appendix~\ref{app:method} specifies the interface computation, the three
training objectives, and the reconstruction-replay schedule.
Appendix~\ref{app:existing-method-audit} defines the Matched, Deranged, and
Receiver-only conditions, the donor-selection rules, and the adapter-only
control for every audited method, including the numerical checks used to verify
that the interventions enter each method through its original injection path.
Appendix~\ref{app:experimental-setup} documents model checkpoints, layer
assignments, training data and per-stage hyperparameters, prompt templates and
decoding settings, evaluation splits and scoring rules, and the evidence
partition used in the Private protocol. Appendix~\ref{app:ablations} reports
the ablation protocol together with the run metadata of each variant. An
anonymized artifact containing the interface implementation, the trained
interface weights, and the evaluation and intervention scripts is available at
\url{https://anonymous.4open.science/r/artifact-review-a8eddb5f-1DDB/}, so that
the reported conditions can be reproduced from the frozen public backbones.


\bibliography{iclr2027_conference,latent_communication}
\bibliographystyle{iclr2027_conference}

\appendix
\input{sections/appendix_toc}
\section{Detailed Method and Training Procedure}
\label{app:method}
\input{sections/appendix_method}

\section{Intervention Protocol and Causal Metrics}
\label{app:intervention}
\subsection{Audit Settings for Existing Latent Communication Methods}
\label{app:existing-method-audit}

Figure~\ref{fig:information-use-gap}(a) evaluates each method under three
paired conditions on the same benchmark examples. \emph{Matched} injects the latent
message generated by the sharer for the evaluated question. \emph{Deranged}
keeps the receiver input and communication interface fixed but replaces that
message with one generated for a different question from the same benchmark.
\emph{Receiver-only} omits the transferred message. Consequently,
$A_M-A_D$ measures the value of the correct question--message pairing while
the interface remains active, whereas $A_M-A_R$ measures the complete system's
improvement over the native receiver.

For C2C in panels~(a) and~(b), we use Qwen2.5-0.5B-Instruct as the
sharer and Qwen3-0.6B as the receiver on MMLU-Redux, OpenBookQA, and
ARC-Challenge. Panel~(b) draws its Matched and adapter-only accuracies
from the same reverse-pair measurements as
Figure~\ref{fig:message-intervention}(b).
The intervention replaces the projected sharer cache supplied to the receiver.
The adapter-only condition averages fused-cache outputs produced from deranged
sharer messages for each receiver input and evaluates the resulting
receiver-conditioned cache; $A_{\mathrm{adapter}}$ denotes its accuracy. For LatentMAS, we apply the same
question-level reassignment to the transferred latent message in its Qwen3-4B
configuration on GSM8K and ARC-Challenge.

All values in Figure~\ref{fig:information-use-gap} are aggregate accuracy point
estimates on the corresponding evaluation split. Multiple-choice benchmarks use
option accuracy, while GSM8K uses final-answer accuracy. Panel~(c) evaluates
MMLU-Redux: the C2C series fixes Qwen2.5-0.5B-Instruct as the receiver and varies
Qwen3 sharers from 0.6B to 8B; the See What I See series fixes Qwen3-4B as the
receiver and compares 8B and 14B sharers. Sharer-only values follow the standalone
MMLU-Redux protocol used in our main results table.

\input{sections/appendix_intervention}
\input{sections/appendix_adapter_only}

\section{Experimental Setup and Reproducibility}
\label{app:experimental-setup}
\input{sections/appendix_setup}

\section{Additional Ablations and Robustness}
\label{app:ablations}
\input{sections/appendix_ablation}

\section{Sharer Scaling Analysis}
\label{app:sharer-scaling}
\input{sections/appendix_scaling}

\section{Receiver Use of the Draft}
\label{app:transmitted-info}
\input{sections/appendix_cases}

\section{Baseline Reproduction and Implementation Checks}
\label{app:baseline-repro}
\input{sections/appendix_baseline}

\section{Case Study}
\label{app:case-examples}
\input{sections/appendix_case_study}

\end{document}

%% file: math_commands.tex
\usepackage{amsmath,amsfonts,bm}

\def\eqref#1{equation~\ref{#1}}

\def\1{\bm{1}}

\DeclareMathAlphabet{\mathsfit}{\encodingdefault}{\sfdefault}{m}{sl}
\SetMathAlphabet{\mathsfit}{bold}{\encodingdefault}{\sfdefault}{bx}{n}



%% file: sections/introduction.tex
Model collaboration matters when models can exchange what they know and have
computed \citep{tran2025multiagentcollaboration,guo2024llmmultiagentsurvey}.
Text requires one model to decode its computation and another to re-encode it
\citep{wu2024autogen,hong2024metagpt,zou2025latentmas}; latent communication
instead passes internal states directly, whether as embedding-level signals
\citep{pham2024ciphers}, hidden-state trajectories
\citep{ramesh2025communicating,du2025latentspace,tang2025statedelta,zheng2025thoughtcommunication},
or key--value caches
\citep{vaswani2017attention,liu2024droidspeak,shi2025kvcomm,jin2026agentprimitives,li2026lesslatent}.
This is especially appealing between heterogeneous models with complementary
capabilities or information
\citep{fu2025c2c,chen2026seewhatisee,feinashley2025mixtureofthoughts}. The
question is therefore not whether a model can describe its reasoning, but
whether a lightweight interface can make its internal computation genuinely
useful to another.

Existing latent interfaces report clear end-task gains for the receiver
\citep{fu2025c2c,zou2025latentmas,chen2026seewhatisee}. Such gains establish
that the trained system is useful, but not that the receiver benefits from the
states it was given: the same improvement would arise if the receiver merely
exploited the adaptation introduced by training the interface. Separating these
explanations requires a contrast that changes the message while holding the
interface fixed. Writing $A_M$, $A_D$, and $A_R$ for
receiver accuracy under a \emph{Matched} message produced for the current
question, a \emph{Deranged} message produced for a different question, and a
\emph{Receiver-only} condition without communication, we separate
\begin{equation}
    G = A_M - A_R, \qquad P = A_M - A_D,
    \label{eq:intro-contrasts}
\end{equation}
where the \emph{system gain} $G$ values the complete system and the
\emph{pairing gain} $P$ values correctly paired content. Section~\ref{sec:information-use-gap}
shows double-digit system gains with pairing gains below one point. We call this
the \emph{information-use gap}: content-independent gains make the sharer
dispensable and cannot convey the advantage of a stronger partner.

\noindent\emph{Can a lightweight latent interface make a receiver benefit
specifically from what the current sharer computed?}

Answering affirmatively requires two properties at once. The message must be
\textbf{informative}: the transmitted states must contain task-relevant
computation the sharer formed for the current input, not a generic summary of
its parameters. And the receiver must make \textbf{effective use} of it: its
answers have to change in a way that depends on receiving the correctly paired
message, not merely on the presence of a trained interface. Meeting both removes
the two failure modes above: a large pairing gain certifies that the sharer's
contribution cannot be absorbed into a receiver-conditioned component, and a
channel that pays for content lets a stronger sharer deliver more.

\textbf{Draft-KV} supplies both. For an informative message, the
sharer first drafts an answer and sends the key--value states formed during that
computation, rather than a static prompt encoding
\citep{hao2024coconut,zhu2025latentreasoning}. A lightweight gated bridge maps
these states into the frozen receiver's attention while both models remain
frozen \citep{hu2022lora,chen2026seewhatisee}. For effective use, progressive
training \citep{bengio2009curriculum} first establishes cross-model readability,
then supervises answers from the draft under a guard that limits harm from a
mismatched message.

\noindent\textbf{Contributions.}
\begin{itemize}\setlength{\itemsep}{1pt}\setlength{\topsep}{2pt}
    \item \textbf{Criterion.} The pairing gain separates content-dependent
    communication from system improvement, and exposes an information-use gap
    in existing latent interfaces (Section~\ref{sec:information-use-gap}).
    \item \textbf{Method.} Draft-KV sends draft-position key--value states
    through a progressively trained $1.05$M-parameter gated interface,
    $348\times$ smaller than C2C~\citep{fu2025c2c} and $36\times$ smaller than
    DLC~\citep{chen2026seewhatisee} (Section~\ref{sec:method}).
    \item \textbf{Evidence.} Gains depend on paired content in all 35 Public
    settings; with a Qwen3-8B sharer, MMLU-Redux
    \citep{hendrycks2021mmlu,gema2025mmluredux} reaches $78.04\%$ Matched,
    $36.40\%$ Deranged, and $37.45\%$ Receiver-only. Gains grow with sharer
    capability and transfer to held-out tasks (Section~\ref{sec:experiments}).
\end{itemize}

%% file: sections/related_work.tex
\paragraph{Language-model collaboration.}
Language-model collaboration combines complementary computations through
structured interaction. AutoGen supports configurable agent conversations
\citep{wu2024autogen}, and MetaGPT organizes specialized roles through
standardized workflows \citep{hong2024metagpt}. Multiagent debate improves
answers through repeated exchanges of proposed solutions and reasoning
\citep{du2023debate}, while Mixture-of-Agents aggregates responses across
successive layers of models \citep{wang2024mixtureofagents}. Optima trains
communication policies to balance task performance, token efficiency, and
readability \citep{chen2025optima}. These methods establish the value of
coordination and communication training, mostly through text, whereas we study
the interface itself: what one model must send for another to benefit from the
computation it just performed.

\paragraph{Embeddings and hidden states.}
CIPHER communicates soft embeddings derived from vocabulary distributions,
retaining alternatives discarded by token sampling \citep{pham2024ciphers}.
\citet{ramesh2025communicating} combine intermediate activations across agents,
while State Delta Encoding augments text with token-wise state-transition
trajectories \citep{tang2025statedelta}. Interlat transmits last-layer hidden
states and learns to compress latent messages \citep{du2025latentspace}.
Thought Communication identifies shared and private latent factors underlying
agent states and uses their sharing structure to organize communication
\citep{zheng2025thoughtcommunication}. Mixture of Thoughts routes queries among
frozen heterogeneous experts and combines their hidden states through learned
cross-attention \citep{feinashley2025mixtureofthoughts}. These approaches make
continuous representations an explicit communication medium; our message is the
layerwise KV states produced during a sharer's answer attempt.

\paragraph{KV-cache sharing and relay.}
DroidSpeak reuses caches across same-architecture models, recomputing selected
layers to accelerate prefill \citep{liu2024droidspeak}. KVComm
selects KV pairs by attention-based importance
\citep{shi2025kvcomm}. LatentMAS combines autoregressive latent-thought
generation with shared KV memory for training-free collaboration
\citep{zou2025latentmas}, and Agent Primitives composes reusable reasoning
components whose interactions use KV caches
\citep{jin2026agentprimitives}. Orthogonal BackFill compresses latent relay by
returning a low-rank residual of discarded states to the retained ones
\citep{li2026lesslatent}. These studies address the reuse, organization, and
cost of cache transmission; we ask instead whether the receiver's accuracy
depends on which cache arrives.

\paragraph{Heterogeneous cache alignment.}
Cache-to-Cache (C2C) projects and fuses the sharer's prompt cache into the
receiver's, with gates selecting communication layers
\citep{fu2025c2c}. \citet{chen2026seewhatisee} develop dense latent
communication (DLC) through
positional disentanglement, structured head transformations, receiver-cache
reconstruction, and generation
training. Draft-KV also establishes readability before task use, but
reconstructs \emph{text from messages} rather than matching receiver cache
tensors. Both works transmit the sharer's prompt cache, whereas our message is
formed while the sharer answers and is read as separate memory rather than
fused into the receiver's own.

\paragraph{Latent reasoning and compressed memory.}
Continuous representations also support reasoning and context compression.
Coconut feeds a model's hidden states back as input embeddings for latent
reasoning \citep{hao2024coconut}, and LaViT aligns latent thoughts for
multimodal reasoning \citep{wu2026lavit}. Closest to our setting, SoftCoT learns
a projection from a frozen assistant's instance-specific soft thoughts into a
frozen LLM's embedding space \citep{xu2025softcot}, though it conditions
through input embeddings rather than layerwise attention memory.
Prompt- and context-compression methods learn readable continuous memory before
downstream use \citep{mu2023gist,chevalier2023autocompressors,ge2023icae}, as
our reconstruction stage does across two models rather than within one.

%% file: sections/information_use_gap.tex
\begin{figure}[t]
    \centering
    \includegraphics[width=\linewidth]{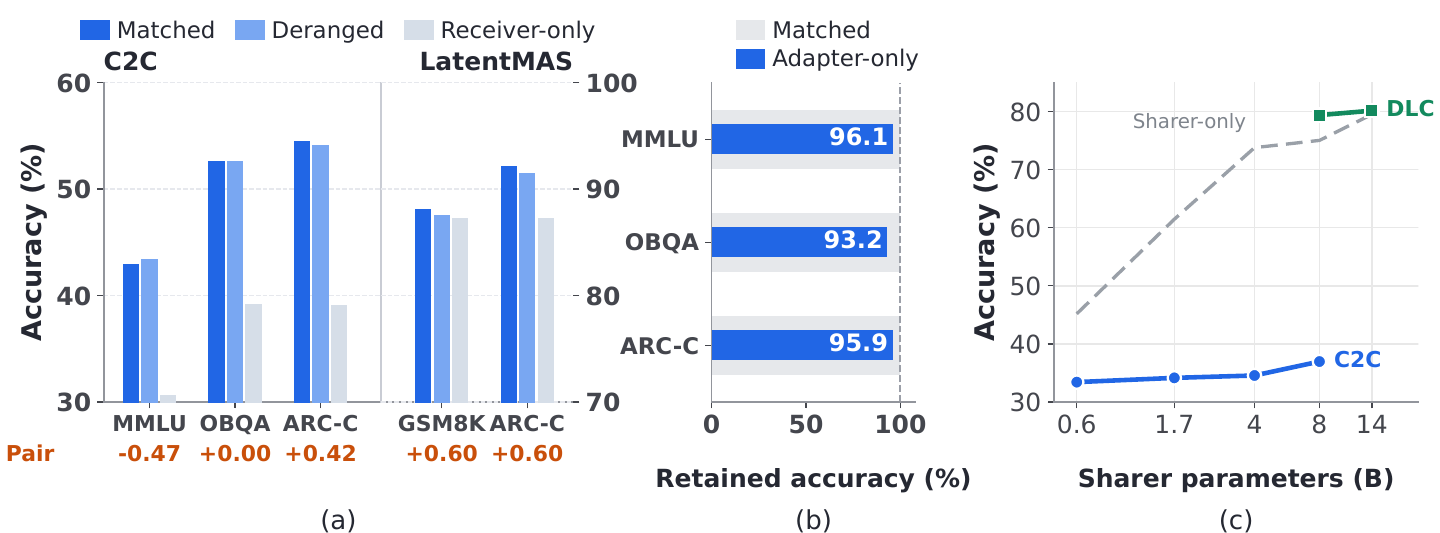}
    \caption{The information-use gap in latent communication.
    (a) Replacing the correctly paired message leaves accuracy essentially
    unchanged in both systems: pairing gain stays within 0.60~pp while system
    gain reaches 15.44~pp (C2C on the left scale, LatentMAS on the right).
    (b) A frozen interface fed mismatched messages still retains 93--96\% of
    Matched accuracy.
    (c) A stronger sharer barely transfers: across C2C's sharers, sharer-only
    accuracy rises by 29.85~pp while system accuracy rises by 3.54~pp.
    All values are aggregate point estimates.}
    \label{fig:information-use-gap}
\end{figure}

Deployed latent interfaces raise three questions: do their gains depend on the
correctly paired message, what sustains the gains that survive it, and
does a stronger sharer produce better answers?
Figure~\ref{fig:information-use-gap} answers them through message replacement,
interface decomposition, and scaling.

\begin{wrapfigure}[12]{l}{0.38\linewidth}
    \centering
    \vspace{-1.3\intextsep}
    \includegraphics[width=\linewidth]{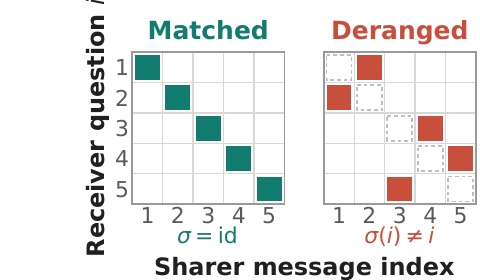}
    \setlength{\abovecaptionskip}{2pt}
    \caption{Message assignment; dashed cells are excluded.}
    \label{fig:matched-deranged}
    \vspace{-1.1\intextsep}
\end{wrapfigure}

We examine learned cache projection in C2C~\citep{fu2025c2c} and cache relay
in LatentMAS~\citep{zou2025latentmas}, and add DLC~\citep{chen2026seewhatisee}
in the scaling analysis below. Within each
method and dataset we
compare the same questions under the Matched, Deranged, and Receiver-only
conditions of Section~\ref{sec:introduction}. Figure~\ref{fig:matched-deranged}
shows the assignment: row $i$ is the receiver's question, the filled
column the message it receives; Appendix~\ref{app:existing-method-audit}
gives the construction. We report the two contrasts of
Equation~\ref{eq:intro-contrasts} in percentage points (pp);
Figure~\ref{fig:information-use-gap}(a) prints $P$ below each dataset as
\emph{Pair}.

\Needspace*{13\baselineskip}
\paragraph{Substantial gains can survive message replacement.}
For C2C, replacing the correctly paired sharer message leaves accuracy almost
unchanged across all three benchmarks in
Figure~\ref{fig:information-use-gap}(a), with Qwen2.5-0.5B-Instruct sharing
to Qwen3-0.6B. On MMLU-Redux, a system gain of 12.29~pp accompanies a pairing
gain of $-0.47$~pp. On OpenBookQA and ARC-Challenge, system gains reach
13.40~pp and 15.44~pp, while pairing gains are 0.00~pp and 0.42~pp,
respectively.
LatentMAS keeps a positive pairing gain on both of its benchmarks, but only
0.60~pp of its 4.86~pp system gain on ARC-Challenge depends on the pairing.
Across the five method--dataset pairs we examine, pairing gain never exceeds
0.60~pp, whereas system gain reaches 15.44~pp. The C2C row behind
Figure~\ref{fig:information-use-gap}(a) is the authors' released fuser and is a
point estimate. To ask whether its near-zero pairing gain is specific to that
one released checkpoint, we trained C2C fusers ourselves under the authors'
recipe---three training checkpoints of the
Qwen2.5-0.5B-Instruct$\to$Qwen3-0.6B direction and one
Llama-3.2-3B-Instruct$\to$Qwen2.5-0.5B-Instruct
fuser---whose 95\% per-question paired bootstrap intervals on the
Matched--Deranged difference contain zero in
13 of 20 cases across the five Public benchmarks, the seven exceptions staying
within a few points of zero (Table~\ref{tab:c2c-bootstrap}); implementation
checks appear in Appendix~\ref{app:baseline-repro}.

\paragraph{Sources of the gain.}
\label{sec:interface-adaptation}
C2C's adapter-only evaluation helps locate the retained improvement. Averaging
fused-cache outputs produced under deranged sharer messages isolates a
receiver-conditioned adapter component, preserving interface adaptation without
the correctly paired message. We report the \emph{retained accuracy}
$\rho_{\mathrm{adapter}}=A_{\mathrm{adapter}}/A_M$, the share of Matched
accuracy that survives.

For the Qwen2.5-0.5B-Instruct to Qwen3-0.6B pair used in panel~(a),
adapter-only retains 96.06\% of Matched accuracy on MMLU-Redux, 95.93\%
on ARC-Challenge, and 93.16\% on OpenBookQA in
Figure~\ref{fig:information-use-gap}(b).
Matched exceeds adapter-only by 1.69, 2.22, and 3.60~pp respectively, so both
interventions leave C2C close to its Matched accuracy. Communication training
should make the supplied information useful beyond what the interface already
provides.

\paragraph{Sharer scaling.}
The practical value of communication also depends on how well the receiver
benefits from a more capable sharer. Figure~\ref{fig:information-use-gap}(c)
compares sharer-only and system accuracy on MMLU-Redux while fixing the
receiver within each method. Across C2C's 0.6B--8B sharers, standalone
sharer accuracy rises by 29.85~pp, while system accuracy rises by 3.54~pp.
With the Qwen3-4B receiver used by \citet{chen2026seewhatisee}, moving from
an 8B to a 14B sharer raises standalone sharer accuracy by 4.48~pp but changes
system accuracy by only 0.77~pp.

A channel whose contribution is largely fixed by the interface cannot carry
much more when the sharer knows more, so most of the stronger partner's
advantage stays stranded on its own side of the interface.

\begin{takeaway}
Existing latent interfaces can improve receiver accuracy with little benefit
from correctly paired content, while stronger sharers do not consistently
improve the system.
\end{takeaway}

%% file: sections/method.tex
\begingroup
\setlength{\abovedisplayskip}{5.5pt}
\setlength{\belowdisplayskip}{5.5pt}
\setlength{\abovedisplayshortskip}{3pt}
\setlength{\belowdisplayshortskip}{3pt}

Draft-KV builds the message from the sharer's attempt at answering. Under
causal attention, prompt-position key--value states are unchanged by the answer
that follows, whereas draft-position states incorporate the question together
with the preceding answer tokens (Figure~\ref{fig:method-overview}).
\begin{figure}[t]
    \centering
    \includegraphics[width=\linewidth]{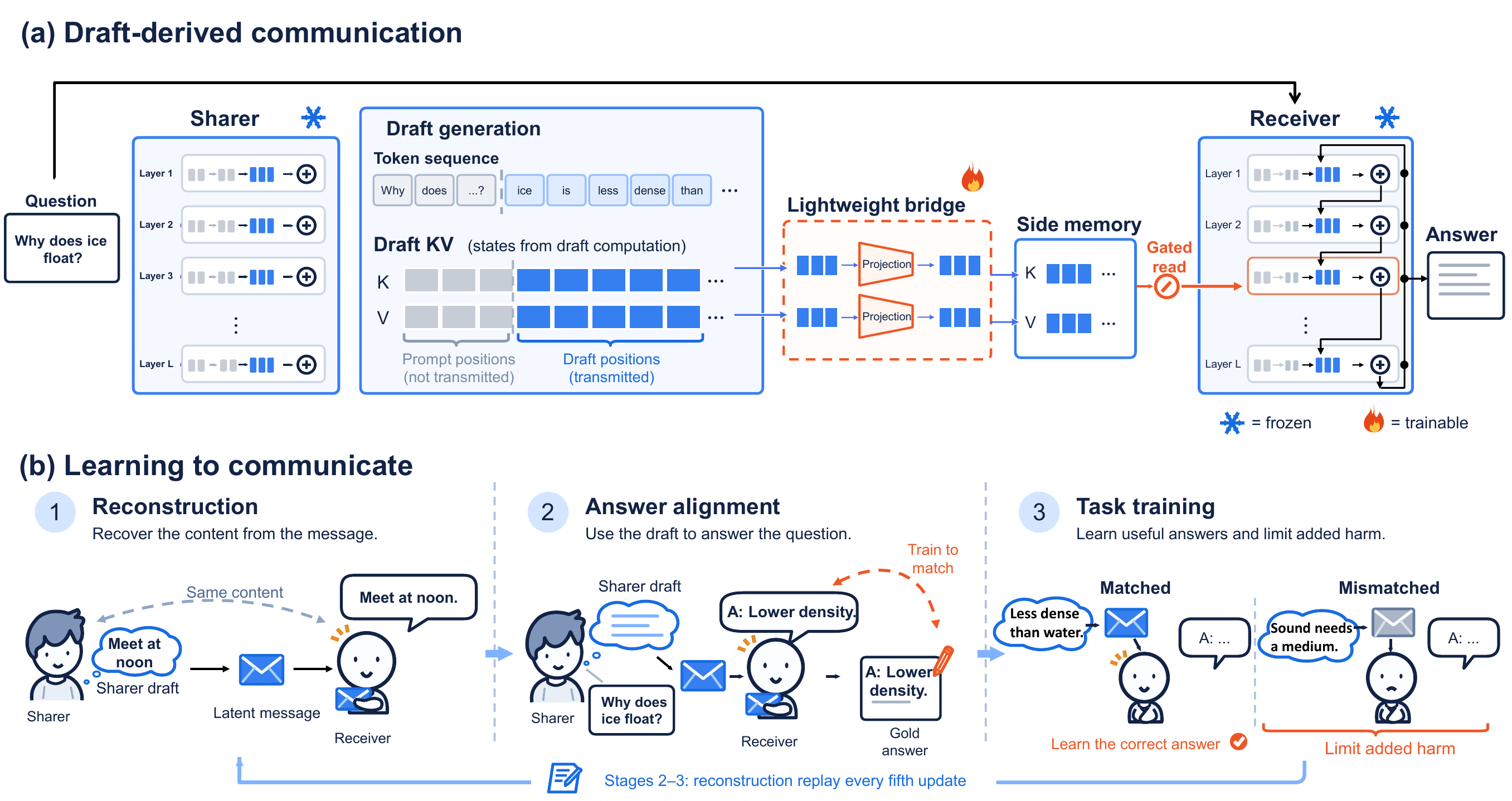}
    \caption{Draft-KV. (a) The frozen sharer drafts an answer, and the
    key--value states formed over prompt and draft are projected into the
    receiver's layout to form the packet $E_\theta(x,d)$, which the frozen
    receiver reads through a gated branch that reuses its own projections.
    (b) The same parameters $\theta$ pass through three training stages.}
    \label{fig:method-overview}
\end{figure}

A frozen sharer $\mathcal{S}_\phi$ assists a frozen receiver $\mathcal{R}_\psi$
on an input $x$, which the receiver must answer with
$y=(y_1,\dots,y_T)$ in its own vocabulary. The trainable parameters are the
communication weights $\theta=\{W_K^{\ell},W_V^{\ell},a^{\ell}\}_{\ell\in\mathcal{I}_R}$,
where $\mathcal{I}_R$ indexes the receiver layers that receive communication and
$\pi(\ell)$ assigns a sharer layer to each. Because the two models
communicate through key--value features, their tokenizers, sequence lengths,
hidden sizes, and KV-head counts may differ
(Appendix~\ref{app:method-shapes}).

\subsection{Draft-Derived Key--Value Messages}
\label{sec:method-message}

The sharer greedily decodes a draft $d=(d_1,\dots,d_L)$ from $x$ until a stop
token or length cap, without seeing the dataset answer. Fixing those
tokens, we run the frozen sharer once over $[x;d]$ and collect its key--value
tensors $K_S^{s},V_S^{s}$ at each layer $s=\pi(\ell)$; a visibility mask $b$
exposes only draft positions.

\subsection{Cross-Model Projection and Gated Injection}
\label{sec:method-interface}

Sharer and receiver states live in different spaces, so each communication layer
$\ell$ flattens the sharer's KV heads at layer $s=\pi(\ell)$ and applies two
bias-free linear maps, one for keys and one for values,
\begin{equation}
    \widetilde{K}^{\ell}=K_S^{s}W_K^{\ell},
    \qquad
    \widetilde{V}^{\ell}=V_S^{s}W_V^{\ell},
    \label{eq:projection}
\end{equation}
where each row is a message token and the outputs are reshaped into the
receiver's KV-head layout. Together with the visibility mask, these form the
packet $E_\theta(x,d)=\{(\widetilde{K}^{\ell},\widetilde{V}^{\ell},b)\}_{\ell\in\mathcal{I}_R}$.

The receiver reads the packet through an attention branch that reuses its own
frozen query and output projections, normalization, and head layout, as shown
in Figure~\ref{fig:method-overview}(a); the branch adds no
parameters beyond one signed gate per KV head,
$g_k^{\ell}=\tanh(a_k^{\ell})$. Writing $\Delta H^{\ell}$ for its gated output,
the branch joins the native self-attention in the same residual stream,
\begin{equation}
    \overline{H}^{\ell}=H^{\ell}+\mathrm{SelfAttn}^{\ell}(U^{\ell})+\Delta H^{\ell},
    \label{eq:injection}
\end{equation}
where $U^{\ell}$ denotes the layer's input-normalized states. The packet stays a
separate memory rather than being concatenated into the receiver's
self-attention cache, and the gates are initialized to zero.
Appendix~\ref{app:method-attention} gives the attention weights, masking
convention, gate placement, and the gates' gradient paths.
Because the branch borrows every other weight from the receiver, the interface
size follows the two models' cache widths, not their depth or
parameter count (Appendix~\ref{app:method-parameters}).

\subsection{Progressive Training}
\label{sec:method-training}

Readability and use are different skills; training both at once gives the
receiver a message it cannot yet read. We therefore train in three stages, each
inheriting $\theta$ from the last
(Figure~\ref{fig:method-overview}(b)). All stages use the same
teacher-forced loss, length-normalized per example,
\begin{equation}
    N_i(E)=-\frac{1}{T_i}\sum_{t=1}^{T_i}
    \log p_\theta\big(y_{i,t}\mid x_i,y_{i,<t},E\big),
    \label{eq:token-loss}
\end{equation}
normalized over the receiver's full vocabulary, excluding prompts and padding
and supervising terminal tokens. The stages differ in what the packet carries
and what the receiver must produce, with per-stage loss reductions in
Appendix~\ref{app:method-losses}.

\paragraph{Stage 1: Message reconstruction.}
\label{sec:stage1}
We take the last assistant turn of an OpenHermes conversation as a payload
$m_i$, append a per-sample transmission key so that the payload contains
sample-specific content, and have the sharer encode $[c_i;m_i]$ under teacher
forcing. The receiver sees only a fixed decoding instruction
$u_{\mathrm{rec}}$---not the conversation, the question, or the key---and is
trained to reproduce $m_i$ from the packet alone by minimizing
$\mathcal{L}_{\mathrm{rec}}$, the token-weighted batch reduction of Equation~\ref{eq:token-loss} with
$x_i=u_{\mathrm{rec}}$ and $y_i=m_i$.

\paragraph{Stage 2: Answer alignment.}
The packet now comes from the sharer's own draft, $E_i=E_\theta(x_i,d_i)$, and
the target is the dataset reference answer $y_i$ for the same conversation,
giving $\mathcal{L}_{\mathrm{ans}}$. Because the draft may be wrong and the
reference is not fed to the sharer, the receiver learns to answer with the draft
states rather than copy them, while every fifth update replays
$\mathcal{L}_{\mathrm{rec}}$ to keep the code decodable.

\paragraph{Stage 3: Answer-text training with a one-sided guard.}
\label{sec:stage3}
On tasks with candidate options, the sharer receives the question and options
and is asked to explain briefly and end with its chosen option in full; the gold
index never enters its input. The receiver is asked for the answer content alone
and is supervised on the full gold answer text, not on an option label. Within
the training split we fix a derangement $\sigma$ with $\sigma(i)\neq i$, and
compare three packets on identical receiver inputs and answer prefixes: Matched
sends the sample's own packet, Deranged sends $E_\theta(x_{\sigma(i)},d_{\sigma(i)})$,
and Receiver-only sends nothing. Writing $N_i^{M}$, $N_i^{D}$, and $N_i^{R}$ for
the resulting losses, the stage minimizes
\begin{equation}
    \mathcal{L}^{(3)}=\frac{1}{B}\sum_{i=1}^{B}
    \Big(N_i^{M}
    +\lambda_p\big[N_i^{D}-\mathrm{sg}(N_i^{R})-\tau\big]_+\Big),
    \qquad \lambda_p=0.1,\ \tau=0.1,
    \label{eq:stage3}
\end{equation}
where $[\cdot]_+=\max(0,\cdot)$ and $\mathrm{sg}$ stops gradient. The first term
raises the likelihood of the gold answer under the correctly paired message. The
second is a one-sided guard: it activates only when a mismatched packet makes
the gold answer more than $\tau$ nats per token costlier than sending nothing,
then pushes that damage down, so the receiver does not learn to be
misled by whatever arrives. Stage 3 keeps the one-in-five reconstruction replay
(Appendix~\ref{app:method-replay}). At inference the sharer drafts, the
interface projects those states, and the receiver decodes with the
packet fixed and available throughout.
\endgroup

%% file: sections/experiments_setup.tex
\paragraph{Training dataset.}
Stages~1 and~2 use OpenHermes conversations for message reconstruction and
answer alignment, respectively. Stage~3 uses ARC-Easy and ARC-Challenge
training data~\citep{clark2018arc}; OpenHermes reconstruction examples are
replayed every fifth update in Stages~2 and~3. ARC test and the other evaluation benchmarks are
held out from Stage~3 training and checkpoint selection. Each model pair is
trained separately, with both language models frozen
(Appendix~\ref{app:method}).

\paragraph{Evaluation settings.}
We evaluate under two protocols. In the \emph{Public} protocol, sharer and
receiver see the same question. We use five multiple-choice
benchmarks: MMLU-Redux for general
knowledge~\citep{gema2025mmluredux}, ARC-Easy and ARC-Challenge for science
questions~\citep{clark2018arc}, OpenBookQA for fact-based
reasoning~\citep{mihaylov2018openbookqa}, and C-EVAL for Chinese
knowledge~\citep{huang2023ceval}. In the \emph{Private} protocol, we split the
annotated gold evidence of HotpotQA and 2WikiMultihopQA at random between them,
so each model holds information absent from the other's input.
We use Qwen3 models from 0.6B to 8B~\citep{yang2025qwen3},
Qwen2.5-0.5B-Instruct~\citep{qwen2025qwen25}, and Llama-3.2-3B-Instruct. Pairing these
models tests scaling across parameter sizes, heterogeneous communication, and
reversed sharer--receiver roles; the specific configurations appear with their
results. We compare against
\emph{Sharer-only} (S-only), which scores the answer extracted from the
sharer's generated draft; \emph{Receiver-only} (R-only), which disables
communication; \emph{Text-to-Text} (T2T), which supplies the same draft as
text to the receiver; and \emph{Cache-to-Cache} (C2C)~\citep{fu2025c2c}, whose
released fuser covers one of the seven model pairs of
Table~\ref{tab:main-results} and whose remaining six rows we retrained
under the authors' recipe (Appendix~\ref{app:baseline-repro}).
On multiple-choice benchmarks, receiver-based methods select the
option with the highest first-token logit and we report accuracy; the
Private benchmarks are scored by exact match. Matched and Deranged
interventions follow Section~\ref{sec:introduction}, and
Appendix~\ref{app:experimental-setup} gives checkpoints, interface sizes,
hyperparameters, prompts, and evaluation splits.

%% file: sections/main_results_table.tex
\begin{table}[t!]
\centering
\caption{Main results across model pairs, benchmarks, and evaluation protocols. Public settings are scored by option accuracy, Private settings by exact match. Cache-to-Cache and Draft-KV entries show Matched/{\scriptsize Deranged} scores, where the Deranged message comes from a different question. \textbf{Bold} and \underline{underlined} values denote the best and second-best score in each row. Model names omit the \texttt{-Instruct} suffix of Qwen2.5-0.5B-Instruct and Llama-3.2-3B-Instruct; Qwen3 names are as released (Appendix~\ref{app:setup-models}).}
\label{tab:main-results}
\begingroup
\definecolor{ProtocolPublic}{HTML}{E4EDF4}
\definecolor{ProtocolPrivate}{HTML}{F5E8DA}
\fontsize{7.4}{8.2}\selectfont
\setlength{\tabcolsep}{2.2pt}
\renewcommand{\arraystretch}{1.0}
\begin{tabular}{@{}lllrrrrr@{}}
\toprule
\multicolumn{3}{c}{Evaluation setting} & \multicolumn{5}{c}{Score (\%)} \\
\cmidrule(lr){1-3}\cmidrule(lr){4-8}
Model Pair & Benchmark & Protocol & Receiver & Sharer & Text-to-Text & Cache-to-Cache & Draft-KV \\
\midrule
\multirow{7}{*}{\shortstack[l]{Qwen3-0.6B\\$\rightarrow$ Qwen2.5-0.5B}} & MMLU-Redux & \cellcolor{ProtocolPublic} & 37.45 & \underline{45.19} & 42.56 & 33.43\,/{\fontsize{6.4}{7.2}\selectfont 33.40} & \textbf{46.11}\,/{\fontsize{6.4}{7.2}\selectfont 36.59} \\
 & ARC-Easy & \cellcolor{ProtocolPublic} & 63.47 & \underline{69.15} & 66.71 & 56.31\,/{\fontsize{6.4}{7.2}\selectfont 53.11} & \textbf{74.28}\,/{\fontsize{6.4}{7.2}\selectfont 61.91} \\
 & ARC-Challenge & \cellcolor{ProtocolPublic} & 40.10 & \underline{51.71} & 49.49 & 39.51\,/{\fontsize{6.4}{7.2}\selectfont 35.32} & \textbf{54.52}\,/{\fontsize{6.4}{7.2}\selectfont 39.76} \\
 & OpenBookQA & \cellcolor{ProtocolPublic} & 43.40 & 44.60 & \underline{45.60} & 41.00\,/{\fontsize{6.4}{7.2}\selectfont 40.00} & \textbf{52.00}\,/{\fontsize{6.4}{7.2}\selectfont 41.20} \\
 & C-EVAL & \cellcolor{ProtocolPublic}\multirow{-5}{*}{Public} & \underline{41.75} & 41.38 & 41.46 & 37.74\,/{\fontsize{6.4}{7.2}\selectfont 36.11} & \textbf{47.40}\,/{\fontsize{6.4}{7.2}\selectfont 40.56} \\
 & HotpotQA & \cellcolor{ProtocolPrivate} & \textbf{32.28} & 11.93 & 13.20 & N/A & \underline{21.91}\,/{\fontsize{6.4}{7.2}\selectfont 14.47} \\
 & 2WikiMultihopQA & \cellcolor{ProtocolPrivate}\multirow{-2}{*}{Private} & \textbf{26.50} & 8.30 & 11.40 & N/A & \underline{17.50}\,/{\fontsize{6.4}{7.2}\selectfont 16.90} \\
\midrule
\multirow{7}{*}{\shortstack[l]{Qwen3-1.7B\\$\rightarrow$ Qwen2.5-0.5B}} & MMLU-Redux & \cellcolor{ProtocolPublic} & 37.45 & \underline{61.47} & 60.19 & 34.16\,/{\fontsize{6.4}{7.2}\selectfont 32.69} & \textbf{65.07}\,/{\fontsize{6.4}{7.2}\selectfont 36.84} \\
 & ARC-Easy & \cellcolor{ProtocolPublic} & 63.47 & \underline{89.52} & 88.26 & 57.37\,/{\fontsize{6.4}{7.2}\selectfont 50.59} & \textbf{92.89}\,/{\fontsize{6.4}{7.2}\selectfont 61.66} \\
 & ARC-Challenge & \cellcolor{ProtocolPublic} & 40.10 & \underline{79.44} & 75.85 & 36.77\,/{\fontsize{6.4}{7.2}\selectfont 33.36} & \textbf{83.53}\,/{\fontsize{6.4}{7.2}\selectfont 37.63} \\
 & OpenBookQA & \cellcolor{ProtocolPublic} & 43.40 & \underline{71.00} & 65.40 & 41.40\,/{\fontsize{6.4}{7.2}\selectfont 37.00} & \textbf{76.00}\,/{\fontsize{6.4}{7.2}\selectfont 42.40} \\
 & C-EVAL & \cellcolor{ProtocolPublic}\multirow{-5}{*}{Public} & 41.75 & \underline{55.79} & 53.71 & 34.77\,/{\fontsize{6.4}{7.2}\selectfont 34.18} & \textbf{61.52}\,/{\fontsize{6.4}{7.2}\selectfont 40.42} \\
 & HotpotQA & \cellcolor{ProtocolPrivate} & \underline{32.28} & 21.18 & 22.68 & N/A & \textbf{35.04}\,/{\fontsize{6.4}{7.2}\selectfont 12.37} \\
 & 2WikiMultihopQA & \cellcolor{ProtocolPrivate}\multirow{-2}{*}{Private} & \textbf{26.50} & 12.80 & 14.90 & N/A & \underline{19.10}\,/{\fontsize{6.4}{7.2}\selectfont 13.50} \\
\midrule
\multirow{7}{*}{\shortstack[l]{Qwen3-4B\\$\rightarrow$ Qwen2.5-0.5B}} & MMLU-Redux & \cellcolor{ProtocolPublic} & 37.45 & \underline{73.79} & 71.15 & 34.57\,/{\fontsize{6.4}{7.2}\selectfont 34.34} & \textbf{75.98}\,/{\fontsize{6.4}{7.2}\selectfont 25.80} \\
 & ARC-Easy & \cellcolor{ProtocolPublic} & 63.47 & \underline{96.42} & 93.31 & 57.45\,/{\fontsize{6.4}{7.2}\selectfont 56.31} & \textbf{96.63}\,/{\fontsize{6.4}{7.2}\selectfont 61.03} \\
 & ARC-Challenge & \cellcolor{ProtocolPublic} & 40.10 & \textbf{91.98} & 84.98 & 38.48\,/{\fontsize{6.4}{7.2}\selectfont 37.54} & \underline{91.81}\,/{\fontsize{6.4}{7.2}\selectfont 39.16} \\
 & OpenBookQA & \cellcolor{ProtocolPublic} & 43.40 & \underline{85.40} & 77.80 & 39.60\,/{\fontsize{6.4}{7.2}\selectfont 40.00} & \textbf{86.40}\,/{\fontsize{6.4}{7.2}\selectfont 42.00} \\
 & C-EVAL & \cellcolor{ProtocolPublic}\multirow{-5}{*}{Public} & 41.75 & \underline{69.99} & 63.67 & 37.44\,/{\fontsize{6.4}{7.2}\selectfont 36.40} & \textbf{71.03}\,/{\fontsize{6.4}{7.2}\selectfont 41.31} \\
 & HotpotQA & \cellcolor{ProtocolPrivate} & \underline{32.28} & 29.04 & 28.57 & N/A & \textbf{40.34}\,/{\fontsize{6.4}{7.2}\selectfont 14.76} \\
 & 2WikiMultihopQA & \cellcolor{ProtocolPrivate}\multirow{-2}{*}{Private} & 26.50 & 29.00 & \underline{29.50} & N/A & \textbf{37.20}\,/{\fontsize{6.4}{7.2}\selectfont 17.70} \\
\midrule
\multirow{7}{*}{\shortstack[l]{Qwen3-8B\\$\rightarrow$ Qwen2.5-0.5B}} & MMLU-Redux & \cellcolor{ProtocolPublic} & 37.45 & \underline{75.04} & 72.05 & 36.97\,/{\fontsize{6.4}{7.2}\selectfont 36.45} & \textbf{78.04}\,/{\fontsize{6.4}{7.2}\selectfont 36.40} \\
 & ARC-Easy & \cellcolor{ProtocolPublic} & 63.47 & \underline{97.56} & 93.52 & 58.16\,/{\fontsize{6.4}{7.2}\selectfont 55.72} & \textbf{98.02}\,/{\fontsize{6.4}{7.2}\selectfont 62.63} \\
 & ARC-Challenge & \cellcolor{ProtocolPublic} & 40.10 & \underline{92.83} & 84.73 & 40.53\,/{\fontsize{6.4}{7.2}\selectfont 39.42} & \textbf{93.77}\,/{\fontsize{6.4}{7.2}\selectfont 38.57} \\
 & OpenBookQA & \cellcolor{ProtocolPublic} & 43.40 & \underline{92.60} & 80.80 & 41.20\,/{\fontsize{6.4}{7.2}\selectfont 39.00} & \textbf{92.80}\,/{\fontsize{6.4}{7.2}\selectfont 43.00} \\
 & C-EVAL & \cellcolor{ProtocolPublic}\multirow{-5}{*}{Public} & 41.75 & \underline{69.84} & 65.82 & 37.96\,/{\fontsize{6.4}{7.2}\selectfont 36.03} & \textbf{74.81}\,/{\fontsize{6.4}{7.2}\selectfont 39.67} \\
 & HotpotQA & \cellcolor{ProtocolPrivate} & 32.28 & \underline{35.21} & 34.59 & N/A & \textbf{42.48}\,/{\fontsize{6.4}{7.2}\selectfont 15.05} \\
 & 2WikiMultihopQA & \cellcolor{ProtocolPrivate}\multirow{-2}{*}{Private} & 26.50 & \underline{37.60} & \underline{37.60} & N/A & \textbf{39.50}\,/{\fontsize{6.4}{7.2}\selectfont 9.10} \\
\midrule
\multirow{7}{*}{\shortstack[l]{Llama-3.2-3B\\$\rightarrow$ Qwen2.5-0.5B}} & MMLU-Redux & \cellcolor{ProtocolPublic} & 37.45 & \underline{64.44} & 62.96 & 31.18\,/{\fontsize{6.4}{7.2}\selectfont 30.11} & \textbf{64.67}\,/{\fontsize{6.4}{7.2}\selectfont 35.90} \\
 & ARC-Easy & \cellcolor{ProtocolPublic} & 63.47 & 88.68 & \textbf{89.44} & 49.28\,/{\fontsize{6.4}{7.2}\selectfont 47.01} & \underline{89.18}\,/{\fontsize{6.4}{7.2}\selectfont 63.13} \\
 & ARC-Challenge & \cellcolor{ProtocolPublic} & 40.10 & \underline{79.18} & 75.34 & 33.53\,/{\fontsize{6.4}{7.2}\selectfont 32.42} & \textbf{79.61}\,/{\fontsize{6.4}{7.2}\selectfont 37.97} \\
 & OpenBookQA & \cellcolor{ProtocolPublic} & 43.40 & \underline{79.40} & 73.60 & 38.40\,/{\fontsize{6.4}{7.2}\selectfont 35.60} & \textbf{80.40}\,/{\fontsize{6.4}{7.2}\selectfont 42.20} \\
 & C-EVAL & \cellcolor{ProtocolPublic}\multirow{-5}{*}{Public} & 41.75 & \underline{43.24} & 42.79 & 33.06\,/{\fontsize{6.4}{7.2}\selectfont 33.21} & \textbf{47.33}\,/{\fontsize{6.4}{7.2}\selectfont 41.16} \\
 & HotpotQA & \cellcolor{ProtocolPrivate} & \textbf{32.28} & 27.74 & 28.19 & N/A & \underline{28.54}\,/{\fontsize{6.4}{7.2}\selectfont 14.20} \\
 & 2WikiMultihopQA & \cellcolor{ProtocolPrivate}\multirow{-2}{*}{Private} & \textbf{26.50} & 22.20 & 22.00 & N/A & \underline{25.70}\,/{\fontsize{6.4}{7.2}\selectfont 17.00} \\
\midrule
\multirow{7}{*}{\shortstack[l]{Qwen2.5-0.5B\\$\rightarrow$ Qwen3-0.6B}} & MMLU-Redux & \cellcolor{ProtocolPublic} & 30.63 & 36.36 & 39.99 & \underline{42.92}\,/{\fontsize{6.4}{7.2}\selectfont 43.39} & \textbf{45.01}\,/{\fontsize{6.4}{7.2}\selectfont 34.11} \\
 & ARC-Easy & \cellcolor{ProtocolPublic} & 59.51 & 54.42 & 56.94 & \underline{72.47}\,/{\fontsize{6.4}{7.2}\selectfont 72.56} & \textbf{73.02}\,/{\fontsize{6.4}{7.2}\selectfont 58.46} \\
 & ARC-Challenge & \cellcolor{ProtocolPublic} & 39.08 & 39.76 & 50.26 & \underline{54.52}\,/{\fontsize{6.4}{7.2}\selectfont 54.10} & \textbf{55.38}\,/{\fontsize{6.4}{7.2}\selectfont 38.23} \\
 & OpenBookQA & \cellcolor{ProtocolPublic} & 39.20 & 40.20 & 46.00 & \underline{52.60}\,/{\fontsize{6.4}{7.2}\selectfont 52.60} & \textbf{53.20}\,/{\fontsize{6.4}{7.2}\selectfont 37.60} \\
 & C-EVAL & \cellcolor{ProtocolPublic}\multirow{-5}{*}{Public} & 31.05 & 33.21 & 33.06 & \textbf{41.75}\,/{\fontsize{6.4}{7.2}\selectfont 41.01} & \underline{41.16}\,/{\fontsize{6.4}{7.2}\selectfont 30.68} \\
 & HotpotQA & \cellcolor{ProtocolPrivate} & 11.93 & \textbf{32.28} & 19.84 & N/A & \underline{31.65}\,/{\fontsize{6.4}{7.2}\selectfont 17.44} \\
 & 2WikiMultihopQA & \cellcolor{ProtocolPrivate}\multirow{-2}{*}{Private} & \textbf{28.00} & 10.00 & 14.50 & N/A & \underline{23.50}\,/{\fontsize{6.4}{7.2}\selectfont 24.40} \\
\midrule
\multirow{7}{*}{\shortstack[l]{Qwen3-8B\\$\rightarrow$ Qwen3-4B}} & MMLU-Redux & \cellcolor{ProtocolPublic} & 71.64 & 75.04 & \underline{78.96} & 71.00\,/{\fontsize{6.4}{7.2}\selectfont 70.70} & \textbf{79.94}\,/{\fontsize{6.4}{7.2}\selectfont 70.44} \\
 & ARC-Easy & \cellcolor{ProtocolPublic} & 94.49 & 97.56 & \textbf{98.19} & 94.40\,/{\fontsize{6.4}{7.2}\selectfont 94.53} & \underline{98.11}\,/{\fontsize{6.4}{7.2}\selectfont 93.90} \\
 & ARC-Challenge & \cellcolor{ProtocolPublic} & 87.29 & \underline{92.83} & 91.30 & 86.69\,/{\fontsize{6.4}{7.2}\selectfont 87.03} & \textbf{93.77}\,/{\fontsize{6.4}{7.2}\selectfont 85.49} \\
 & OpenBookQA & \cellcolor{ProtocolPublic} & 78.80 & \textbf{92.60} & 88.40 & 78.60\,/{\fontsize{6.4}{7.2}\selectfont 78.20} & \underline{92.20}\,/{\fontsize{6.4}{7.2}\selectfont 76.40} \\
 & C-EVAL & \cellcolor{ProtocolPublic}\multirow{-5}{*}{Public} & 70.43 & 69.84 & \underline{72.51} & 67.76\,/{\fontsize{6.4}{7.2}\selectfont 67.90} & \textbf{75.78}\,/{\fontsize{6.4}{7.2}\selectfont 69.47} \\
 & HotpotQA & \cellcolor{ProtocolPrivate} & \underline{42.71} & 35.21 & 41.72 & N/A & \textbf{45.12}\,/{\fontsize{6.4}{7.2}\selectfont 37.88} \\
 & 2WikiMultihopQA & \cellcolor{ProtocolPrivate}\multirow{-2}{*}{Private} & 40.90 & 37.60 & \underline{44.10} & N/A & \textbf{45.00}\,/{\fontsize{6.4}{7.2}\selectfont 39.00} \\
\bottomrule
\end{tabular}
\endgroup
\end{table}

%% file: sections/main_results.tex
\paragraph{Public context.}
Table~\ref{tab:main-results} shows that Draft-KV consistently improves the
receiver under shared input: it beats the receiver alone in all 35
settings, is most accurate in 30, and exceeds T2T in 33, with
an average gain of 5.85 points. The improvement holds across
sharer scales, model families, and the reversed configuration.
Draft-KV and T2T communicate the same generated draft through different
representations, so their gap indicates that draft-derived states preserve
information the decoded text does not fully retain. Draft-KV can also surpass
the sharer: on ARC-Challenge it recovers 24.1\% of the questions the
sharer answers incorrectly while losing only 1.1\% of those it answers
correctly, so the receiver does more than relay the sharer's answer
(Appendix~\ref{app:case-counts}).

\paragraph{Private context.}
Across HotpotQA and 2WikiMultihopQA, Draft-KV beats T2T in all 14
model--benchmark settings, by 6.41 points on average. It ranks first in seven
settings and second in the other seven, and where it ranks first it also
exceeds both models answering independently, so collaboration can improve on
either model's local prediction when sharer and receiver hold different
evidence. Here the packet is the receiver's only route to the sharer's half of
the evidence, so a deranged packet substitutes another question's evidence and
falls 10.98~pp below Receiver-only, against 1.41~pp under Public. Pairing gain
is positive in 13 of the 14 settings.

%% file: sections/message_intervention.tex
\begin{figure}[!ht]
    \centering
    \begin{minipage}[t]{0.635\linewidth}
        \centering
        \includegraphics[width=\linewidth]{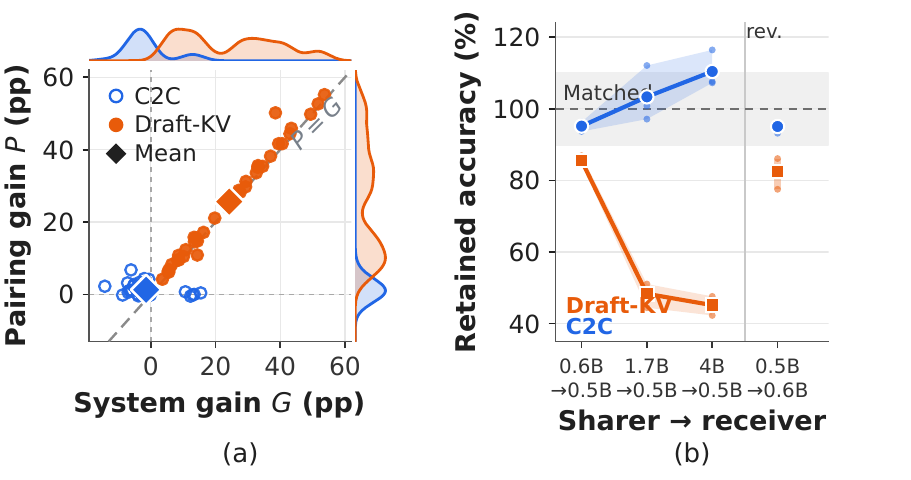}
        \caption{Message interventions and adapter-only comparisons under the
        Public protocol. In (b), lines and bands give the mean and min--max of
        retained accuracy over the three benchmarks, dots the individual
        benchmarks, and the dashed line and shaded strip mark Matched accuracy
        and a $\pm10$~pp margin.}
        \label{fig:message-intervention}
    \end{minipage}\hfill
    \begin{minipage}[t]{0.345\linewidth}
        \centering
        \includegraphics[width=\linewidth]{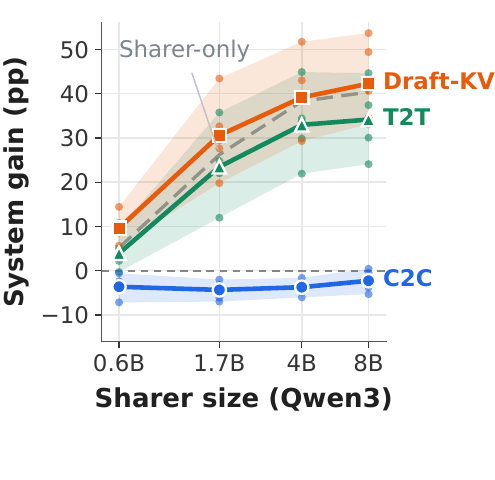}
        \caption{Sharer scaling with a fixed Qwen2.5-0.5B-Instruct receiver.
        Bands give the range across the five benchmarks.}
        \label{fig:sharer-scaling}
    \end{minipage}
\end{figure}

\paragraph{Pairing gain and system gain.}
Draft-KV addresses the information-use gap of
Section~\ref{sec:information-use-gap}: its improvement over the receiver
depends on the correctly paired message. Figure~\ref{fig:message-intervention}(a)
shows positive system gain $G=A_M-A_R$ and pairing gain $P=A_M-A_D$
in all 35 Public settings, with diamonds marking the method means of
24.24 and 25.65~pp, respectively. The dashed diagonal marks $P=G$, where
$A_D=A_R$ and a deranged message costs exactly the gain. Draft-KV clusters
there: the mismatch harm $A_R-A_D$ has a median of 1.19~pp and stays within
2.5~pp in 33 of the 35 settings, reaching 11.65~pp on MMLU-Redux with the
Qwen3-4B sharer (Table~\ref{tab:intervention-distribution}). C2C instead
collapses onto the origin: its pairing gain averages 1.31~pp and stays
within $\pm7$~pp everywhere, and its
system gain is negative in 29 of the 35 settings. With $P=G+(A_R-A_D)$, a
1.41~pp mean harm against a 24.24~pp mean system gain makes the pairing gain
matched-answer improvement rather than mismatch damage
(Appendix~\ref{app:guard-ablation}).

\paragraph{Adapter-only comparison.}
We extend the adapter-only analysis of Section~\ref{sec:information-use-gap}
to both methods across four model pairs on MMLU-Redux, ARC-Challenge, and
OpenBookQA, averaging each frozen interface's output over five mismatched
donors (Appendix~\ref{app:adapter-only}), and plot the retained accuracy
$\rho_{\mathrm{adapter}}=A_{\mathrm{adapter}}/A_M$ in
Figure~\ref{fig:message-intervention}(b). C2C retains
93.16--116.42\% of Matched accuracy across its 12 settings, whereas Draft-KV
with the 1.7B and 4B sharers retains 42.28--51.05\% and falls below
Receiver-only in all six. Holding the query convention fixed and varying only
the donor, restoring the correct one is worth 26.96 and 29.49~pp, so the drop
follows the donor content: the Matched gains of the stronger sharers do not
survive donor averaging through the trained interface.

%% file: sections/sharer_scaling.tex
Draft-KV turns stronger sharers into larger receiver gains.
With Qwen2.5-0.5B-Instruct fixed as the receiver, we scale Qwen3 sharers from
0.6B to 8B. In Figure~\ref{fig:sharer-scaling}, solid lines give the mean system
gain $G=A_M-A_R$ over the five benchmarks and the dashed line the sharer-only
reference, the gain the sharer's accuracy represents over the receiver.
Draft-KV's system gain rises at every size step on all five benchmarks,
addressing the scaling gap of Section~\ref{sec:information-use-gap}, and beats
that reference in 19 of 20 settings. Its mean system gain rises by 32.6~pp from
0.6B to 8B, against 30.2~pp for T2T and 1.4~pp
for C2C, and it stays more accurate than T2T at every size. What grows is the
capability behind the draft rather than its length, which does not order these
gains (Appendix~\ref{app:sharer-scaling}).

%% file: sections/ablation_study.tex
\begin{table}[!htbp]
\centering
\caption{Message-source and training-stage ablations for Qwen3-1.7B
$\rightarrow$ Qwen2.5-0.5B-Instruct under the Public protocol.}
\label{tab:ablation}
\footnotesize
\setlength{\tabcolsep}{4pt}
\begin{tabular}{@{}lrrrr@{}}
\toprule
& \multicolumn{2}{c}{ARC-Challenge} & \multicolumn{2}{c}{MMLU-Redux} \\
\cmidrule(lr){2-3}\cmidrule(lr){4-5}
Variant & Matched & Pairing gain & Matched & Pairing gain \\
\midrule
Full Draft-KV & \textbf{83.53} & 45.90 & \textbf{65.07} & 28.23 \\
Prompt KV & 49.83 & 0.17 & 43.59 & 0.09 \\
w/o reconstruction pretraining & 76.71 & 38.74 & 52.40 & 16.58 \\
w/o answer alignment & 80.03 & 41.72 & 55.40 & 19.25 \\
\bottomrule
\end{tabular}
\par\smallskip
\begin{minipage}{\linewidth}
\footnotesize
Matched is $A_M$ (\%); pairing gain is $P=A_M-A_D$ (pp). Receiver-only accuracy
is $40.10\%$ on ARC-C (test split) and $37.45\%$ on MMLU-Redux. Stage deletions
retain replay in the remaining stages (Appendix~\ref{app:ablation-core}).
\end{minipage}
\end{table}

With a fixed Qwen3-1.7B sharer and Qwen2.5-0.5B-Instruct receiver, we ablate
message source and training stages on ARC-C and MMLU-Redux
(Table~\ref{tab:ablation}). Replacing draft-position KV with prompt-position KV
reduces Matched accuracy by 33.70 and 21.48~pp and nearly eliminates pairing
gain, which falls to 0.17 and 0.09~pp---the level at which existing interfaces
operate, C2C averaging $1.31$~pp in Section~\ref{sec:message-intervention}.
This reproduces the information-use gap inside our architecture, showing that
draft-position states make the pairing matter.
Deleting either stage lowers Matched accuracy and pairing gain on both
benchmarks, and the losses are markedly larger on MMLU-Redux, which no stage
trains on: Stage~3 updates partly compensate for a weaker cross-model mapping
on the task they optimize, and the held-out benchmark is where that
compensation runs out. Appendix~\ref{app:ablations} adds per-benchmark metrics
and the guard analysis.

%% file: sections/appendix_toc.tex
\definecolor{tocred}{HTML}{B22222}

\newcommand{\apptocsec}[3]{%
  \vspace{9pt}\noindent
  {\bfseries\color{tocred}%
    \makebox[1.8em][l]{#1}#2\unskip\ \nobreak\hfill\pageref{#3}}\par}

\newcommand{\apptocsub}[3]{%
  \vspace{2pt}\noindent\hspace{2.2em}%
  \makebox[2.4em][l]{#1}#2\nobreak\ \dotfill\ \pageref{#3}\par}

\clearpage
\begin{center}
  {\LARGE\bfseries Appendix of Draft-KV}
\end{center}
\vspace{1em}

\noindent\textsc{Contents}
\vspace{0.6em}

\apptocsec{A}{Detailed Method and Training Procedure}{app:method}
\apptocsub{A.1}{Shapes and Message Extraction}{app:method-shapes}
\apptocsub{A.2}{External Attention and Gated Injection}{app:method-attention}
\apptocsub{A.3}{Interface Size}{app:method-parameters}
\apptocsub{A.4}{Stage Objectives and Replay}{app:method-losses}

\apptocsec{B}{Intervention Protocol and Causal Metrics}{app:intervention}
\apptocsub{B.1}{Audit Settings for Existing Latent Communication Methods}{app:existing-method-audit}
\apptocsub{B.2}{Message Interventions in Our Experiments}{app:our-intervention}
\apptocsub{B.3}{Adapter-Only Evaluation}{app:adapter-only}

\apptocsec{C}{Experimental Setup and Reproducibility}{app:experimental-setup}
\apptocsub{C.1}{Models and Interface Configurations}{app:setup-models}
\apptocsub{C.2}{Training Data and Stage Configuration}{app:setup-training}
\apptocsub{C.3}{Prompts and Decoding}{app:setup-prompts}
\apptocsub{C.4}{Evaluation Data and Scoring}{app:setup-eval}
\apptocsub{C.5}{Compute and Software}{app:setup-compute}

\apptocsec{D}{Additional Ablations and Robustness}{app:ablations}
\apptocsub{D.1}{Core Ablations: Results and Protocol}{app:ablation-core}
\apptocsub{D.2}{Guard Ablation: Mismatch Harm versus Pairing Gain}{app:guard-ablation}

\apptocsec{E}{Sharer Scaling Analysis}{app:sharer-scaling}

\apptocsec{F}{Receiver Use of the Draft}{app:transmitted-info}
\apptocsub{F.1}{Disagreement with the Sharer}{app:case-counts}

\apptocsec{G}{Baseline Reproduction and Implementation Checks}{app:baseline-repro}

\apptocsec{H}{Case Study}{app:case-examples}
\apptocsub{H.1}{Corrected Cases}{app:case-corrected}
\apptocsub{H.2}{Medium Cases}{app:case-medium}
\apptocsub{H.3}{Mismatch Cases}{app:case-mismatch}
\apptocsub{H.4}{Lost Cases}{app:case-lost}
\apptocsub{H.5}{Split-Evidence Generation Cases}{app:case-split}

\clearpage

%% file: sections/appendix_method.tex
This appendix completes the interface computation and the training objectives of
Section~\ref{sec:method}. Model checkpoints, layer assignments, and per-pair
interface sizes are given in Appendix~\ref{app:experimental-setup}.

\subsection{Shapes and Message Extraction}
\label{app:method-shapes}

We use row vectors for token features. Let $\ell\in\mathcal I_R$ be a receiver
communication layer and $s=\pi(\ell)$ its assigned sharer layer under the fixed
map $\pi$. Write $H_S^s,D_S^s$ and $H_R^\ell,D_R^\ell$ for the number of KV
heads and the head width at those layers, $C_S^s=H_S^sD_S^s$ and
$C_R^\ell=H_R^\ell D_R^\ell$ for the flattened KV widths, and
$d_{\mathrm{model},R}^\ell$ for the width of the receiver's residual stream.
With $n_S$ sharer positions in the extracted sequence and $n_R$ receiver
positions in a forward pass,
\begin{equation}
    K_S^s,V_S^s\in\mathbb R^{n_S\times H_S^s\times D_S^s},
    \qquad
    \widetilde K^\ell,\widetilde V^\ell
        \in\mathbb R^{n_S\times H_R^\ell\times D_R^\ell},
    \qquad
    H^\ell,U^\ell\in\mathbb R^{n_R\times d_{\mathrm{model},R}^\ell}.
    \label{eq:app-packet-shapes}
\end{equation}
The two lengths are independent and each model keeps its own tokenizer, so no
alignment between their token positions is required. When the two models receive
different instructions for the same task input $x$, we write $x^S$ and $x^R$.

\paragraph{Extraction point.}
The sharer first decodes $d$ from $x^S$. Its tokens are then held fixed for one
forward pass over $[x^S;d]$, and at each selected layer $s$ we extract keys
\emph{after} the sharer's native key normalization, where present, and
\emph{before} rotary position encoding; values come from the same layer and
positions. The sharer's own attention still uses its native rotary encoding and
causal mask, so taking pre-rotary keys removes position information from the
transmitted tensors, not from the computation that contextualized them. In
Stage~1 the same interface processes a supplied assistant payload $m$ after its
conversation context instead of a generated draft.

\paragraph{Visibility.}
A single mask covers the transmitted payload in either case,
\begin{equation}
b_j =
\begin{cases}
1 & \text{if \(j\) is a transmitted draft or reconstruction-payload position,}\\[2pt]
0 & \text{for context or padding positions.}
\end{cases}
\label{eq:app-visibility}
\end{equation}
The packet may retain tensors over the full sharer sequence while exposing only
the positions with $b_j=1$, and context information still reaches the receiver
through the contextualized payload states. The mask is shared across
communication layers for a given message.

\paragraph{Projection.}
For each position $j$, flattening runs over the two feature axes rather than the
sequence axis, and the two bias-free maps
$W_K^\ell,W_V^\ell\in\mathbb R^{C_S^{\pi(\ell)}\times C_R^\ell}$ give
\begin{equation}
    \widetilde k_j^\ell=\operatorname{vec}(K_{S,j}^{\pi(\ell)})W_K^\ell,
    \qquad
    \widetilde v_j^\ell=\operatorname{vec}(V_{S,j}^{\pi(\ell)})W_V^\ell,
    \label{eq:app-projection-vectors}
\end{equation}
whose outputs reshape into $H_R^\ell\times D_R^\ell$. A layer's matrices are
shared across all message positions, while different communication layers have
separate matrices; because the maps are dense over the flattened width, they can
mix information across sharer KV heads before producing the receiver's head
layout.

\subsection{External Attention and Gated Injection}
\label{app:method-attention}

At receiver layer $\ell$, both the native path and the external branch start
from the normalized hidden states
$U^\ell=\operatorname{Norm}_{\mathrm{in}}^\ell(H^\ell)$. Reusing the receiver's
frozen query projection and its query and key normalization, the branch forms
\begin{equation}
    Q^\ell=\operatorname{Norm}_Q^\ell\!\left(
      \operatorname{reshape}(\operatorname{Proj}_Q^\ell(U^\ell))\right),
    \qquad
    \overline K^\ell=\operatorname{Norm}_K^\ell(\widetilde K^\ell),
    \label{eq:app-queries}
\end{equation}
so sharer-side key normalization precedes cross-model projection while
receiver-side key normalization acts on the projected keys in the receiver's
feature space. An absent normalization module is an identity map. No rotary
phase is applied to $Q^\ell$ or $\overline K^\ell$ in the external branch; the
receiver's native self-attention retains its own rotary encoding and causal
mask. Under grouped-query attention, each external KV head is read by the query
heads natively grouped with it, which introduces no parameters.

Writing $\kappa_\ell(h)$ for the KV head assigned to query head $h$, and
masking invisible positions additively, the branch attends over message
positions,
\begin{equation}
    A_{t,\cdot,h}^\ell=\operatorname{softmax}_j\!\left(
        \frac{\langle Q_{t,h}^\ell,
              \overline K_{j,\kappa_\ell(h)}^\ell\rangle}{\sqrt{D_R^\ell}}
        +B_j\right),
    \qquad
    Z_{t,h}^\ell=\sum_{j=1}^{n_S}
        A_{t,j,h}^\ell\widetilde V_{j,\kappa_\ell(h)}^\ell,
    \label{eq:app-external-attention}
\end{equation}
with $B_j=0$ when $b_j=1$ and $-\infty$ otherwise. Every receiver position may
read every visible payload position, since the complete packet exists before
receiver generation starts; no triangular mask relates receiver index $t$ to
message index $j$. The summaries are receiver-conditioned even though the packet
is fixed, because the queries follow the current receiver hidden states.

One scalar gate per receiver KV head, $g_k^\ell=\tanh(a_k^\ell)$, scales each
head's summary before concatenation and the receiver's frozen output projection
$W_O^\ell$,
\begin{equation}
    \Delta H_t^\ell=
    \operatorname{Concat}_{h}
       \left(g_{\kappa_\ell(h)}^\ell Z_{t,h}^\ell\right)W_O^\ell,
    \qquad
    \overline H^\ell=H^\ell+\operatorname{SelfAttn}^\ell(U^\ell)+\Delta H^\ell,
    \label{eq:app-gated-output}
\end{equation}
after which the layer's native MLP proceeds unchanged and layers outside
$\mathcal I_R$ use $\Delta H^\ell=0$. All query heads mapped to one KV head
share its gate, which is more flexible than a single shared scalar gate on the
whole projected output. The external packet stays separate from the native
self-attention cache:
during inference the former is fixed while the latter grows with decoding.

\paragraph{Initialization and gradient paths.}
\label{app:method-gates}
The projection matrices use Xavier initialization and the gates start at zero,
so $\Delta H^\ell=0$ at initialization and the model computes the native
receiver function; later stages inherit both from the preceding stage. Both
backbones stay frozen and the sharer's draft generation and extraction receive
no gradients, but the receiver computation after injection still carries the
gradient path from the text loss to $\theta$. At exactly zero gates the
projections receive no gradient through the branch while the gates do, and once
the gates move away from zero the same objective updates the projections.

\subsection{Interface Size}
\label{app:method-parameters}

Per communication layer, the two dense maps contribute $2C_S^{\pi(\ell)}C_R^\ell$
parameters and the gate vector $H_R^\ell$; reshaping, masking, the layer
assignment, and the head mapping add none, and the reused receiver projections
and normalization modules belong to the frozen backbone. Hence
\begin{equation}
    |\theta|=\sum_{\ell\in\mathcal I_R}
        \left(2C_S^{\pi(\ell)}C_R^\ell+H_R^\ell\right).
    \label{eq:app-parameter-count}
\end{equation}
The size is therefore set by the two models' cache widths and the number of
communication layers, not by their depth or parameter count, and message length
does not enter it at all. Appendix~\ref{app:experimental-setup} lists the
resulting counts for every model pair we train.

\subsection{Stage Objectives and Replay}
\label{app:method-losses}

All stages supervise the receiver's full-vocabulary token probabilities under
teacher forcing. For a target of $T_i$ supervised tokens, including its terminal
token and excluding prompt and padding positions,
\begin{equation}
    N_i(E)=-\frac{1}{T_i}\sum_{t=1}^{T_i}
        \log p_\theta(y_{i,t}\mid x_i^R,y_{i,<t},E),
    \label{eq:app-token-nll}
\end{equation}
measured in nats per target token. The teacher-forced receiver prefix never
enters the sharer's draft-generation input.

\paragraph{Stages 1 and 2.}
Both reduce this loss with token weighting over a microbatch $\mathcal B$, so an
example's total weight is proportional to its supervised length. In Stage~1 the
packet is extracted from $[c_i;m_i]$ and exposes the payload positions, and the
receiver reconstructs the payload $m_i$ from the fixed instruction
$u_{\mathrm{rec}}$, giving $\mathcal L_{\mathrm{rec}}$. In Stage~2 the packet
comes from the sharer's own draft and the target is the reference answer, giving
$\mathcal L_{\mathrm{ans}}$.

\paragraph{Stage 3.}
For each example we build three conditions on the same receiver input, target,
and teacher-forced prefix,
\begin{equation}
    E_i^M=E_\theta(x_i^S,d_i),\qquad
    E_i^D=E_\theta(x_{\sigma(i)}^S,d_{\sigma(i)}),\qquad
    E_i^R=\emptyset,
    \label{eq:app-training-conditions}
\end{equation}
where the fixed derangement $\sigma$ has no self-pairings and the donor
contributes its complete packet, including its visibility mask. Writing
$N_i^c=N_i(E_i^c)$, the stage applies the guard per example and then averages
over examples,
\begin{equation}
    \mathcal L_{\mathrm{task}}^{(3)}(\mathcal B)
        =\frac1B\sum_{i\in\mathcal B}
          \Big(N_i^M+\lambda_p
          \big[N_i^D-\operatorname{sg}(N_i^R)-\tau\big]_+\Big),
        \qquad \lambda_p=0.1,\quad\tau=0.1,
        \label{eq:app-stage3-loss}
\end{equation}
which is $\mathcal L^{(3)}$ in Equation~\ref{eq:stage3}. Two choices here are
deliberate. The matched term gives every example weight $1/B$ regardless of
answer length, unlike the token weighting of the earlier stages. And the hinge
is applied per example rather than to the batch's average loss difference, so
one example's improvement cannot cancel another's excess harm; since $\tau$ is
compared against length-normalized losses, it is a threshold in nats per target
token. The guard is one-sided in effect as well as in form: when active it
lowers the gold-answer NLL under a mismatched packet, and it can never raise
it.

\paragraph{Replay schedule.}
\label{app:method-replay}
Stage~1 optimizes $\mathcal L_{\mathrm{rec}}$ on every update. Stages~2 and~3
alternate four task updates with one reconstruction update, which uses the
Stage~1 payload construction, instruction, and token weighting throughout; the
Stage~3 guard applies to task updates only. This is an alternating schedule
rather than a weighted reconstruction term added to every task loss.
Communication parameters carry over across all three stages, and neither frozen
backbone is reinitialized or optimized.

%% file: sections/appendix_intervention.tex
\subsection{Message Interventions in Our Experiments}
\label{app:our-intervention}

The three conditions reported for Draft-KV throughout
Section~\ref{sec:experiments} share one interface checkpoint, one set of
questions, and one set of cached sharer drafts per configuration. Within a
configuration we change the packet and nothing else: the receiver's prompt,
the gold target, the teacher-forced prefix where applicable, the decoding
settings, and the scoring rule are identical across Matched, Deranged, and
Receiver-only. Matched injects the packet extracted from the evaluated
question's own draft, Deranged injects another question's complete packet
including its visibility mask, and Receiver-only disables the external branch
and is verified to reproduce the native receiver's logits exactly.

\paragraph{Donor pool.}
For each benchmark, the donor pool is exactly the set of examples selected for
that evaluation, after applying the split, subject, and sample-count filters:
5{,}632 for MMLU-Redux, 2{,}376 for ARC-Easy, 1{,}172 for ARC-Challenge, 500 for
OpenBookQA, and 1{,}346 for C-EVAL. Pools are per benchmark, so a donor never
crosses datasets, but within a pool donors are unrestricted by subject or by
evaluation batch. One exception is recorded: for the Qwen3-0.6B to
Qwen2.5-0.5B-Instruct pair, the ARC-Easy and ARC-Challenge test sets are drawn
from a merged pool of 3{,}548 examples, so donors may cross the two ARC splits.

\paragraph{Permutation.}
Donors are assigned by a single global derangement of the pool rather than by
sampling with replacement: we shuffle the example identifiers under a fixed seed
and accept the first shuffle without fixed points, so the assignment is a
bijection in which every example donates exactly once and no example donates to
itself. The mapping is fully determined by the identifier order and the seed. The
main-table evaluations use seed 71031 and the merged-ARC exception uses seed
99173; each run's manifest records full coverage and zero fixed points. Training
uses the same construction inside its own split, so no training derangement can
leak an example's own draft back to it.

\paragraph{Message length and padding.}
Matched and Deranged packets in one target batch share a common tensor width,
and the packet attention mask hides the padded positions, so a donor's original
message length is preserved as a mask rather than by resizing the target batch.
Donor drafts are neither truncated to the target's draft length nor re-decoded.

\paragraph{Distribution of the contrasts.}
Table~\ref{tab:intervention-distribution} reports the system gain, the pairing
gain, and the mismatch harm $A_R-A_D$ over the settings of
Table~\ref{tab:main-results}, separated by protocol. Differences are computed
from the reported two-decimal accuracies.

\begin{table}[htbp]
\centering
\caption{Distribution of Draft-KV intervention contrasts across the settings of
Table~\ref{tab:main-results}. All quantities are in percentage points.}
\label{tab:intervention-distribution}
\small
\setlength{\tabcolsep}{5pt}
\begin{tabular}{@{}lrrrrrrr@{}}
\toprule
 & & \multicolumn{2}{c}{Mean gain} & \multicolumn{4}{c}{Mismatch harm $A_R-A_D$} \\
\cmidrule(lr){3-4}\cmidrule(lr){5-8}
Protocol & $n$ & $G$ & $P$ & Mean & Median & Min & Max \\
\midrule
Public & 35 & 24.24 & 25.65 & 1.41 & 1.19 & $-3.48$ & 11.65 \\
Private & 14 & 2.51 & 13.49 & 10.98 & 11.30 & $-5.51$ & 19.91 \\
\bottomrule
\end{tabular}
\end{table}

%% file: sections/appendix_adapter_only.tex
\subsection{Adapter-Only Evaluation}
\label{app:adapter-only}

\paragraph{Purpose and scope.}
Adapter-only tests how much accuracy a trained communication interface
retains when its output is averaged over messages from other questions.
No parameters are trained: the control reuses the existing checkpoint and
averages its interface output over five donor messages before the receiver
makes one prediction.
What it isolates is a receiver-conditioned component of that interface, which
still depends on the target's receiver input and on which donors were drawn,
rather than a source-free constant or a uniquely identified parameter
contribution.

\paragraph{Data and measurement provenance.}
Table~\ref{tab:adapter-controls} and
Figure~\ref{fig:message-intervention}(b) cover MMLU-Redux, ARC-Challenge,
and OpenBookQA across four model pairs. Receiver-only and Matched
accuracies follow Table~\ref{tab:main-results}; the C2C and Draft-KV
adapter-only accuracies are supplied by the additional evaluations.
All differences and retained accuracies use these main-table references.

\paragraph{MMLU-Redux evaluation details.}
The MMLU-Redux runs use the adapted MMLU-Redux~2.0 test set of
Appendix~\ref{app:setup-eval} ($n=5{,}632$).
The predicted option maximizes the last-position logit of the first token
of its space-prefixed option letter. C2C uses its option-logits prompt;
Draft-KV uses the no-CoT prompt ending in \texttt{The correct answer is}.
Draft-KV reuses saved sharer drafts with a 512-token cap.

\paragraph{MMLU-Redux donor selection.}
Donors come from the target's subject and are grouped by receiver-prompt
token length. For buckets of size at least two, seeds 17, 29, 41, 53, and
67 each produce a permutation without fixed points, providing five donors
$d_1(i),\ldots,d_5(i)$ for target $i$. Donors can repeat across seeds, but
$d_k(i)\neq i$ always holds. For singleton buckets, the other examples
in the same subject are ordered by
$(|L_j-L_i|,L_j,\mathrm{sample\_id}_j)$, where $L_i$ is prompt length.
The nearest five are selected, cycling through candidates if needed.
The evaluator asserts non-self assignment and records exact-length or
fallback selection. The run summary reports approximately 3,634
exact-length and 1,998 fallback targets, with the same donor protocol
for both methods.

\paragraph{One-shot output averaging.}
Let $z_i^{\ell,0}$ be a target-side state from a no-communication forward
pass and $s_j$ a source message. For the frozen interface output map
$\mathcal F_\theta^\ell$, the control injects
\begin{equation}
 B_i^\ell=Q_{\mathrm{bf16}}\!\left(
 \frac15\sum_{k=1}^{5}\operatorname{float}_{32}
 [\mathcal F_\theta^\ell(s_{d_k(i)},z_i^{\ell,0})]\right).
 \label{eq:adapter-donor-average}
\end{equation}
The mean is accumulated in FP32 and cast back to BF16. Target states are
collected once; adapter outputs are not recomputed on a trajectory altered
by communication. Injection uses the method's original insertion point.

\paragraph{C2C: fused-cache output.}
The frozen module is the checkpoint's fuser, including its projection and
gate. For receiver cache $T_i^{\ell,0}$ and source cache $S_j^\ell$, define
\begin{align}
 M_i^\ell&=\mathcal A_\theta^\ell(S_i^\ell,T_i^{\ell,0}),\\
 D_{i,k}^\ell&=\mathcal A_\theta^\ell(S_{d_k(i)}^\ell,T_i^{\ell,0}),\\
 \overline C_i^\ell&=Q_{\mathrm{bf16}}\!\left(
 \frac15\sum_k\operatorname{float}_{32}(D_{i,k}^\ell)\right).
 \label{eq:c2c-adapter-only}
\end{align}
Donor caches are truncated or padded to the target prefix length for batch
alignment. Adapter-only injects $\overline C_i^\ell$ through the same
cache path as Matched. Two FP32 compensation terms, formed by compensated
subtraction, verify reconstruction:
\begin{equation}
 \max\left|Q_{\mathrm{bf16}}\!\left(\overline C_i^\ell+
 C_{\mathrm{hi},i}^\ell+C_{\mathrm{lo},i}^\ell\right)-M_i^\ell\right|
 \leq10^{-6}.
\end{equation}
Adapter-only disables both compensation terms. This verifies numerical
reconstruction through the common path; the algebraic residual is not
itself a causal decomposition of source information.

\paragraph{Draft-KV: residual-update output.}
We load the checkpoint's bias-free K/V projection stack and
per-KV-head gate parameters and verify the loaded communication-state
digest against the checkpoint. Receiver query, normalization, and output
paths remain frozen; evaluation disables gradient recording.
The checkpoint pairs receiver layers $\{14,16,18,20\}$ with sharer layers
$\{18,20,22,24\}$ for Qwen3-\{0.6,1.7,4\}B to Qwen2.5-0.5B-Instruct.
The reversed pair swaps these two layer lists.

Each donor packet is extracted from the frozen sharer's prompt and saved
draft, using unrotated K/V, the trained projections, and a draft-position
visibility mask. Let $\mathcal X_\theta^\ell(U,E)$ be the gated external
attention of Equations~\ref{eq:app-external-attention}--\ref{eq:app-gated-output}.
For one-shot normalized states
$U_i^{\ell,0}=\operatorname{Norm}_{\mathrm{in}}^\ell(H_i^{\ell,0})$,
\begin{align}
 \Delta H_{i,k}^\ell&=\mathcal X_\theta^\ell(U_i^{\ell,0},E_{d_k(i)}),\\
 \overline{\Delta H}_i^\ell&=Q_{\mathrm{bf16}}\!\left(
 \frac15\sum_k\operatorname{float}_{32}(\Delta H_{i,k}^\ell)\right).
 \label{eq:draft-adapter-only}
\end{align}
The frozen receiver query path and key normalization are reused.
GQA repeats KV heads across their query-head groups, and
$\tanh(a_k^\ell)$ gates each KV-head value summary before the output
projection. The update enters the original parallel branch:
\begin{equation}
 \overline H_i^\ell=H_i^\ell+
 \operatorname{SelfAttn}^\ell\!\left(
 \operatorname{Norm}_{\mathrm{in}}^\ell(H_i^\ell)\right)
 +\overline{\Delta H}_i^\ell.
\end{equation}
Native self-attention and the subsequent MLP proceed normally.
Unlike source caches, donor attention outputs already share the target
token dimensions, so no source-length alignment is needed. Neither the
packets nor the receiver logits are averaged.

\input{sections/adapter_only_table}

\paragraph{Retained accuracy.}
Table~\ref{tab:adapter-controls} gives every raw accuracy alongside its
retained ratio $\rho_{\mathrm{adapter}}=A_{\mathrm{adapter}}/A_M$, defined in
Section~\ref{sec:information-use-gap}; these are aggregate point estimates and
do not imply paired statistical significance. Beyond the ranges reported in
Section~\ref{sec:message-intervention}, C2C exceeds $100\%$ in five of its 12
settings, and the two weakest Draft-KV configurations---the 0.6B sharer and the
reversed pair---retain $77.50$--$86.90\%$, so the separation between
adapter-only and Matched grows with sharer capability.

\paragraph{Query convention and self-donor diagnostic.}
Two things change between Matched and adapter-only. Matched forms each query
from the current receiver state, including preceding layers' communication,
whereas adapter-only fixes external queries to the no-communication
trajectory; and the donor packet is no longer the target's own. The C2C
control is verified against the fused cache itself
(Appendix~\ref{app:baseline-repro}); for Draft-KV we separate these two
factors with a self-donor diagnostic, which restores the target's own draft as
the sole donor while keeping the one-shot queries and the averaging path.

Table~\ref{tab:adapter-self-donor} reports the first 512 MMLU-Redux examples.
Holding the query convention fixed, restoring the correct donor is worth
26.96 and 29.49~pp for the 1.7B and 4B sharers. The query convention itself is
close to free: on the same subset the 1.7B sharer reaches 60.55\% under
self-donor against 60.35\% for standard Matched, measured for that pair. The
adapter-only drop therefore follows the donor content rather than the fixed
queries, on a subset that establishes the direction of the effect and not its
exact share of the full Matched--adapter-only gap.

\begin{table}[t]
 \centering
 \caption{Draft-KV self-donor diagnostic on the first 512 MMLU-Redux
 examples. All entries are accuracy (\%). Self-donor and adapter-only
 use the same one-shot queries. All comparisons use this subset.}
 \label{tab:adapter-self-donor}
 \small
 \begin{tabular}{@{}lrrr@{}}
 \toprule
 Sharer $\to$ Receiver & Receiver-only & Adapter-only & Self-donor\\
 \midrule
 Q3-0.6B $\to$ Q2.5-0.5B & 36.13 & 39.65 & 37.89\\
 Q3-1.7B $\to$ Q2.5-0.5B & 36.13 & 33.59 & 60.55\\
 Q3-4B $\to$ Q2.5-0.5B & 36.13 & 34.96 & 64.45\\
 Q2.5-0.5B $\to$ Q3-0.6B & 37.50 & 41.02 & 42.19\\
 \bottomrule
 \end{tabular}
\end{table}

%% file: sections/adapter_only_table.tex
\begin{table}[t]
 \centering
 \caption{Adapter-only comparison across three benchmarks and four model pairs.
 All accuracies and retained accuracies are percentages.
 Receiver-only and Matched follow Table~\ref{tab:main-results};
 only adapter-only accuracies are added from the new evaluations.
 Q3 and Q2.5 denote Qwen3 and Qwen2.5-Instruct.
 Retained accuracy $\rho_{\mathrm{adapter}}=A_{\mathrm{adapter}}/A_M$.
 Values above 100\% mean adapter-only exceeds Matched.}
 \label{tab:adapter-controls}
 \small
 \setlength{\tabcolsep}{3.2pt}
 \begin{tabular}{@{}lrrrrrrr@{}}
 \toprule
 & & \multicolumn{3}{c}{C2C} & \multicolumn{3}{c}{Draft-KV}\\
 \cmidrule(lr){3-5}\cmidrule(l){6-8}
 Sharer $\to$ Receiver & $A_R$ & $A_M$ & $A_{\mathrm{adapter}}$ & $\rho_{\mathrm{adapter}}$
 & $A_M$ & $A_{\mathrm{adapter}}$ & $\rho_{\mathrm{adapter}}$\\
 \midrule
 \multicolumn{8}{c}{\textbf{MMLU-Redux}}\\
 \addlinespace[2pt]
Q3-0.6B $\to$ Q2.5-0.5B & 37.45 & 33.43 & 31.92 & 95.48\% & 46.11 & 40.07 & 86.90\% \\
Q3-1.7B $\to$ Q2.5-0.5B & 37.45 & 34.16 & 34.43 & 100.79\% & 65.07 & 33.22 & 51.05\% \\
Q3-4B $\to$ Q2.5-0.5B & 37.45 & 34.57 & 37.07 & 107.23\% & 75.98 & 36.17 & 47.60\% \\
Q2.5-0.5B $\to$ Q3-0.6B & 30.63 & 42.92 & 41.23 & 96.06\% & 45.01 & 37.64 & 83.63\% \\
 \midrule
 \multicolumn{8}{c}{\textbf{ARC-Challenge}}\\
 \addlinespace[2pt]
Q3-0.6B $\to$ Q2.5-0.5B & 40.10 & 39.51 & 38.05 & 96.30\% & 54.52 & 46.33 & 84.98\% \\
Q3-1.7B $\to$ Q2.5-0.5B & 40.10 & 36.77 & 41.21 & 112.08\% & 83.53 & 37.12 & 44.44\% \\
Q3-4B $\to$ Q2.5-0.5B & 40.10 & 38.48 & 44.80 & 116.42\% & 91.81 & 38.82 & 42.28\% \\
Q2.5-0.5B $\to$ Q3-0.6B & 39.08 & 54.52 & 52.30 & 95.93\% & 55.38 & 42.92 & 77.50\% \\
 \midrule
 \multicolumn{8}{c}{\textbf{OpenBookQA}}\\
 \addlinespace[2pt]
Q3-0.6B $\to$ Q2.5-0.5B & 43.40 & 41.00 & 38.40 & 93.66\% & 52.00 & 44.00 & 84.62\% \\
Q3-1.7B $\to$ Q2.5-0.5B & 43.40 & 41.40 & 40.20 & 97.10\% & 76.00 & 37.60 & 49.47\% \\
Q3-4B $\to$ Q2.5-0.5B & 43.40 & 39.60 & 42.60 & 107.58\% & 86.40 & 39.40 & 45.60\% \\
Q2.5-0.5B $\to$ Q3-0.6B & 39.20 & 52.60 & 49.00 & 93.16\% & 53.20 & 45.80 & 86.09\% \\
 \bottomrule
 \end{tabular}
\end{table}

%% file: sections/appendix_setup.tex
This appendix documents the configurations behind every reported number: the
model pairs and interface sizes, the data and hyperparameters of the three
training stages, the prompts and decoding settings, and the evaluation splits
and scoring rules.

\subsection{Models and Interface Configurations}
\label{app:setup-models}

We use publicly released instruction-tuned and base checkpoints without further
adaptation: \texttt{Qwen/Qwen3-\{0.6B,1.7B,4B,8B\}},
\texttt{Qwen/Qwen2.5-0.5B-Instruct}, and
\texttt{meta-llama/Llama-3.2-3B-Instruct}. Both backbones are frozen in every
run, and an interface is trained separately for each pair.

Table~\ref{tab:setup-interface} lists the layer assignment and the trainable
parameter count per pair. Every pair uses four communication layers. The five
pairs with a Qwen2.5-0.5B-Instruct receiver share the assignment
$\pi$: receiver layers $\{14,16,18,20\}$ read sharer layers $\{18,20,22,24\}$;
the reversed pair swaps the two lists, and the Qwen3-8B to Qwen3-4B pair uses
receiver layers $\{21,24,27,30\}$ with sharer layers $\{23,26,29,32\}$.

Because $|\theta|=\sum_\ell(2C_S^{\pi(\ell)}C_R^\ell+H_R^\ell)$ depends only on
the two models' flattened KV widths, an identical layer assignment does not
imply an identical interface size. Qwen3 and Llama-3.2 sharers all have
$C_S=8\times128=1024$, and Qwen2.5-0.5B-Instruct has $C_R=2\times64=128$, so the
four Qwen3 sharers and the Llama-3.2-3B sharer produce the same $1{,}048{,}584$
parameters against that receiver. This is what the sharer-scaling analysis of
Section~\ref{sec:sharer-scaling} holds fixed. The two pairs with a larger
receiver differ: Qwen3-0.6B as a receiver has eight KV heads, which raises the
gate count from $8$ to $32$, and Qwen3-4B as a receiver has $C_R=1024$, which
raises the projection count eightfold.

\begin{table}[htbp]
\centering
\caption{Interface configuration per model pair. Layer assignments are written
receiver$\leftarrow$sharer. $C_S$ and $C_R$ are flattened KV widths,
$H_R$ the number of receiver KV heads and hence of gates. Model names omit
\texttt{-Instruct} suffixes as in Table~\ref{tab:main-results}.}
\label{tab:setup-interface}
\small
\setlength{\tabcolsep}{4pt}
\begin{tabular}{@{}llrrrrr@{}}
\toprule
Sharer $\to$ Receiver & Layers ($\ell\leftarrow\pi(\ell)$) & $C_S$ & $C_R$ &
$H_R$ & Projections & $|\theta|$ \\
\midrule
Qwen3-0.6B $\to$ Qwen2.5-0.5B & 14,16,18,20 $\leftarrow$ 18,20,22,24 & 1024 & 128 & 2 & 1{,}048{,}576 & 1{,}048{,}584 \\
Qwen3-1.7B $\to$ Qwen2.5-0.5B & 14,16,18,20 $\leftarrow$ 18,20,22,24 & 1024 & 128 & 2 & 1{,}048{,}576 & 1{,}048{,}584 \\
Qwen3-4B $\to$ Qwen2.5-0.5B & 14,16,18,20 $\leftarrow$ 18,20,22,24 & 1024 & 128 & 2 & 1{,}048{,}576 & 1{,}048{,}584 \\
Qwen3-8B $\to$ Qwen2.5-0.5B & 14,16,18,20 $\leftarrow$ 18,20,22,24 & 1024 & 128 & 2 & 1{,}048{,}576 & 1{,}048{,}584 \\
Llama-3.2-3B $\to$ Qwen2.5-0.5B & 14,16,18,20 $\leftarrow$ 18,20,22,24 & 1024 & 128 & 2 & 1{,}048{,}576 & 1{,}048{,}584 \\
Qwen2.5-0.5B $\to$ Qwen3-0.6B & 18,20,22,24 $\leftarrow$ 14,16,18,20 & 128 & 1024 & 8 & 1{,}048{,}576 & 1{,}048{,}608 \\
Qwen3-8B $\to$ Qwen3-4B & 21,24,27,30 $\leftarrow$ 23,26,29,32 & 1024 & 1024 & 8 & 8{,}388{,}608 & 8{,}388{,}640 \\
\bottomrule
\end{tabular}
\end{table}

\subsection{Training Data and Stage Configuration}
\label{app:setup-training}

\paragraph{Reconstruction and answer-alignment data.}
Stages~1 and~2 draw on the first 500{,}000 conversations of OpenHermes~2.5. A
conversation is admitted to Stage~1 if it contains only system, user, and
assistant turns, ends with a non-empty assistant message, and that message is an
exact token suffix of the prompt under both tokenizers. We append a per-sample
transmission key to the payload, deduplicate
on the NFKC-normalized casefolded message, and take conversations in a
seeded shuffle order until the target count is reached. Scanning 132{,}744
conversations yields the 36{,}864 used
per configuration: 32{,}768 for training, 2{,}048 for gate validation, 2{,}048
held in reserve, and 32 for the overfit check below.

Stage~2 reuses the same OpenHermes split with all Stage~1 sample IDs removed. A
record qualifies if its context ends with a user turn and its gold assistant
message is non-empty, after deduplication on the SHA-256 of the
$\{\text{context},\text{gold}\}$ pair; about 280{,}840 records qualify. The
sharer then greedily drafts from the context alone, and we keep only drafts that
terminate in EOS, which leaves roughly 9{,}220 usable rows out of a candidate
pool of 9{,}728--11{,}776. The first 8{,}192 become the training set, followed by
512 for gate validation and 512 held in reserve. No filter in either stage
inspects difficulty, gold correctness, or draft quality.

\paragraph{Stage 3 data.}
Stage~3 uses the ARC-Easy and ARC-Challenge train and validation splits,
4{,}239 examples in total, divided by a fixed stratified split into 3{,}602
training and 637 calibration examples. Sharer drafts are generated offline and
cached before training, together with their prompt tokens, draft mask, decoded
text, and sample identifiers. Reconstruction replay reuses the Stage~1 data:
32{,}768 training conversations and the 2{,}048-conversation gate-validation set.
None of the evaluation benchmarks enters Stage~3 training or checkpoint
selection.

\paragraph{Optimization.}
All stages use AdamW with zero weight decay and gradient clipping at $1.0$, and
separate learning rates for the projections and the gates. Stage~1 begins with a
300-update overfit check on 32 conversations, used only as a go/no-go signal on
the data pipeline; the reported run restarts from a randomly initialized
interface and does not inherit those weights. Stage~1 then trains for 16{,}000
updates at projection learning rate $10^{-3}$ and gate learning rate $10^{-2}$
with an effective batch of 64. Stage~2 trains for 4{,}000 optimizer updates at
$2\times10^{-4}$ and $10^{-3}$ with an effective batch of 32, of which 3{,}200
are answer-alignment updates and 800 are reconstruction replay. Stage~3 trains
for 5{,}000 optimizer updates at $10^{-4}$ and $5\times10^{-4}$ with an
effective batch of 16, of which 4{,}000 are task updates and 1{,}000 are replay,
and uses seed 31847 for every pair. Stages~2 and~3 realize the four-to-one
replay ratio of Appendix~\ref{app:method-replay}; Stage~1 reconstructs on every
update. Update counts and learning rates are identical across pairs; only the
microbatch and accumulation factors differ, to fit the sharer in memory at a
constant effective batch (Table~\ref{tab:setup-stages}).

\begin{table}[htbp]
\centering
\caption{Per-pair microbatch $\times$ gradient accumulation and data seeds for
Stages~1 and~2. Effective batches are 64 and 32 respectively for every pair.
Model names omit \texttt{-Instruct} suffixes as in
Table~\ref{tab:main-results}.}
\label{tab:setup-stages}
\small
\setlength{\tabcolsep}{5pt}
\begin{tabular}{@{}lccccc@{}}
\toprule
& \multicolumn{2}{c}{Stage 1} & \multicolumn{2}{c}{Stage 2} \\
\cmidrule(lr){2-3}\cmidrule(lr){4-5}
Sharer $\to$ Receiver & micro $\times$ accum & seed & micro $\times$ accum & seed \\
\midrule
Qwen3-0.6B $\to$ Qwen2.5-0.5B & $16\times4$ & 20260826 & $4\times8$ & 20260828 \\
Qwen3-1.7B $\to$ Qwen2.5-0.5B & $8\times8$ & 91827 & $4\times8$ & 91827 \\
Qwen3-4B $\to$ Qwen2.5-0.5B & $8\times8$ & 91827 & $4\times8$ & 91827 \\
Qwen3-8B $\to$ Qwen2.5-0.5B & $4\times16$ & 91827 & $4\times8$ & 91827 \\
Llama-3.2-3B $\to$ Qwen2.5-0.5B & $8\times8$ & 91827 & $4\times8$ & 91827 \\
Qwen2.5-0.5B $\to$ Qwen3-0.6B & $16\times4$ & 91827 & $2\times16$ & 91827 \\
Qwen3-8B $\to$ Qwen3-4B & $1\times64$ & 91827 & $1\times32$ & 91827 \\
\bottomrule
\end{tabular}
\end{table}

\paragraph{Length limits.}
In Stage~1 the natural message must occupy at least 16 receiver tokens, the
transmitted message including its key at most 128, the full receiver sequence at
most 256, and the full sharer sequence at most 1{,}024. In Stage~2 the gold
answer spans 16--256 receiver tokens, the receiver context plus gold is capped at
1{,}024, the sharer prompt at 1{,}536, and the sharer prompt plus draft at
3{,}072, with draft decoding capped at 1{,}024 new tokens; the text-to-text
control allows 3{,}072 receiver tokens because it carries the draft as text.
Stage~3 and all reported evaluations cap sharer drafts at 512 new tokens.

\paragraph{Checkpoint selection.}
Every 250 task updates in Stage~3 we evaluate the 637 calibration examples and
the 2{,}048 gate-validation conversations. A snapshot is eligible only if it
preserves the communication acquired earlier: its Matched reconstruction NLL
must satisfy
$\overline N_{\mathrm{rec},M}(\theta)\le1.1\,\overline N_{\mathrm{rec},M}(\theta_2)$
and its reconstruction gap must satisfy
$(\overline N_{\mathrm{rec},R}-\overline N_{\mathrm{rec},M})(\theta)\ge
0.9\,(\overline N_{\mathrm{rec},R}-\overline N_{\mathrm{rec},M})(\theta_2)$,
where $\theta_2$ is the Stage~2 checkpoint. Among eligible snapshots we select
the one with the lowest mean Matched answer NLL on the calibration set. Test
data never participates in this selection.

\begin{figure}[htbp]
    \centering
    \includegraphics[width=\linewidth]{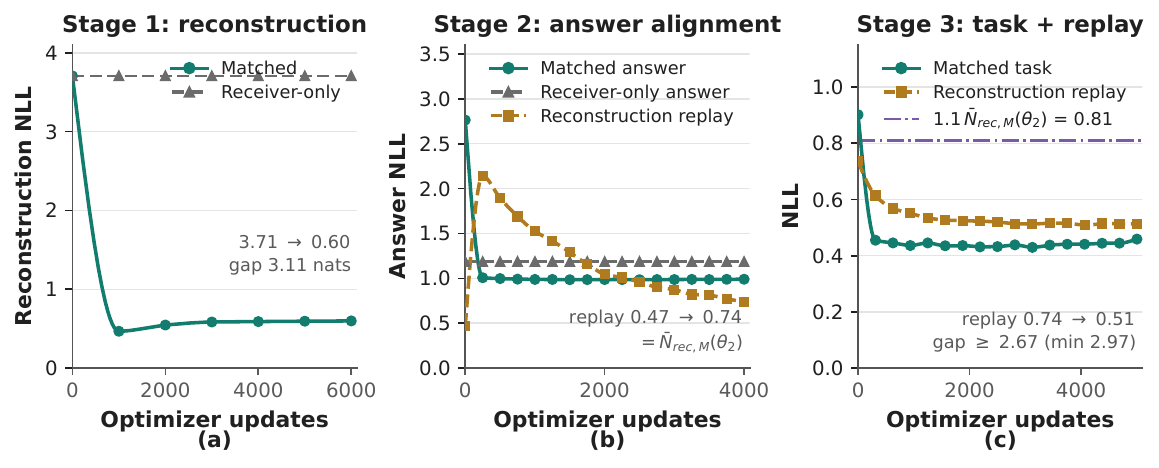}
    \caption{Three-stage training dynamics,
    Qwen3-1.7B$\to$Qwen2.5-0.5B-Instruct, measured on held-out splits.
    (a) Reconstruction NLL, Matched against the constant Receiver-only baseline.
    (b) Matched answer NLL against Receiver-only, with the replayed
    reconstruction NLL that defines $\overline N_{\mathrm{rec},M}(\theta_2)$.
    (c) Matched task NLL with the replayed reconstruction NLL and the
    eligibility bound $1.1\,\overline N_{\mathrm{rec},M}(\theta_2)$. The
    Stage~1 panel shows the first $6{,}000$ of its $16{,}000$ updates, over
    which reconstruction has converged ($3.71\to0.47$ nats by update $1{,}000$
    and $0.60$ at update $6{,}000$). Markers are measurements; curves are
    shape-preserving interpolation between them.}
    \label{fig:three-stage-curves}
\end{figure}

\paragraph{Training dynamics.}
Figure~\ref{fig:three-stage-curves} logs the three stages of this configuration
on held-out splits. Two observations matter for the selection rule above.
First, the replayed reconstruction NLL rises when task supervision starts and
then recovers: to $2.14$ nats by Stage-2 update $250$ and back to $0.74$ at
its end, which is the $\overline N_{\mathrm{rec},M}(\theta_2)$ the rule refers
to. Second, Stage~3 lowers the task loss without giving the code up: Matched
task NLL falls from $0.90$ to $0.46$ while replay stays within
$0.51$--$0.74$, under $1.1\,\overline N_{\mathrm{rec},M}(\theta_2)=0.81$, and
the reconstruction gap stays above $2.97$ nats against the required $2.67$, so
every evaluated snapshot is eligible.

\subsection{Prompts and Decoding}
\label{app:setup-prompts}

Every prompt below is rendered through the model's own chat template. Qwen3
models run with thinking disabled throughout, which for the Qwen3 tokenizer
inserts an empty \texttt{<think></think>} block after the assistant marker.
Sharer drafts are decoded greedily.

\paragraph{Stage 1: reconstruction.}
The receiver sees one fixed instruction for every sample and never sees the
conversation, the question, or the transmission key:
\begin{promptbox}{Stage 1 \textperiodcentered\ Receiver}
[system] You are a lossless communication decoder. Recover the hidden
         message exactly and do not add, remove, explain, or rewrite
         anything.
[user]   Output the hidden message exactly.
\end{promptbox}
\noindent This renders to 45 tokens for the Qwen2.5 receiver and 49 for the
Qwen3 receiver.

\paragraph{Stage 2: answer alignment.}
The sharer receives the OpenHermes conversation with the final gold assistant
turn removed, through the chat template with a generation prompt and no added
instruction, and drafts the next assistant turn. The receiver sees the same
conversation. The text-to-text control instead appends the draft to the
receiver's user turn:
\begin{promptbox}{Stage 2 \textperiodcentered\ Text-to-text receiver}
{original user content}

A collaborating model proposed the draft answer below. Treat it as
fallible evidence, check it, and then answer the original request.
<collaborator_draft>
{sharer draft}
</collaborator_draft>
\end{promptbox}

\paragraph{Stage 3: multiple-choice training.}
The sharer receives the question and all options and is asked for the option
text rather than its letter, so that the draft carries content rather than a
label:
\begin{promptbox}{Stage 3 \textperiodcentered\ Sharer, multiple choice}
Question:
{question}

Options:
A. {choice A}
B. {choice B}
C. {choice C}
D. {choice D}

Explain your reasoning briefly using the provided information. End with
a separate line: Answer: the full text of your chosen option, not its
letter.
\end{promptbox}
\noindent The receiver is supervised on the gold option text under
\begin{promptbox}{Stage 3 \textperiodcentered\ Receiver, multiple choice}
Question:
{question}

Options:
A. {choice A}
B. {choice B}
C. {choice C}
D. {choice D}

Give only the answer text. Do not give an option letter, explanation,
or answer label.
\end{promptbox}

\paragraph{Stage 3: private protocol.}
The sharer prompt inserts its assigned evidence and asks for a short answer;
the receiver prompt drops the option block and keeps the final instruction:
\begin{promptbox}{Stage 3 \textperiodcentered\ Sharer, private protocol}
Question:
{question}

Context:
[{title}]
{text}
...

Explain your reasoning briefly using the provided information. End with
a separate line: Answer: the short answer.
\end{promptbox}

\paragraph{Evaluation.}
Multiple-choice evaluation scores options rather than sampling text, so the
receiver is prompted in the option-selection format used by prior work and its
turn is prefilled with \texttt{The correct answer is}:
\begin{promptbox}{Evaluation \textperiodcentered\ Receiver, option selection}
Accurately answer the following question:

{question}

Choices:
A. {choice A}
B. {choice B}
C. {choice C}
D. {choice D}

Instructions:
- Carefully read the question and all options.
- Select the single most correct answer.
- Respond ONLY in the format "The correct answer is A/B/C/D".
- Do not include explanations or additional text.
\end{promptbox}
\noindent Training therefore supervises answer text while multiple-choice
evaluation reads option logits at a fixed position; the two use the prompts
given above and the same interface checkpoint. Private-protocol evaluation
decodes greedily with at most 128 new tokens. Drafts that reach the 512-token
cap without emitting EOS are truncated and used as they are.

\subsection{Evaluation Data and Scoring}
\label{app:setup-eval}

\paragraph{Benchmarks.}
The Public protocol uses MMLU-Redux~2.0 test ($n=5{,}632$, dataset digest prefix
\texttt{ec464e71}), ARC-Easy test ($n=2{,}376$), ARC-Challenge test
($n=1{,}172$), OpenBookQA test ($n=500$), and C-EVAL validation
($n=1{,}346$). The Private protocol uses HotpotQA ($n=7{,}404$) and
2WikiMultihopQA ($n=1{,}000$) following the split of HippoRAG~2.

\paragraph{Multiple-choice scoring.}
Receiver-based methods predict the option whose space-prefixed letter has the
highest first-token logit at the last prompt position. Questions with fewer
candidates mask the unused letters. Because scoring reads logits rather than
sampling, there are no unparsed outputs. Sharer-only instead extracts the chosen
option from the sharer's decoded draft, which is generated from the same prompt
and under the same 512-token cap as the drafts Draft-KV transmits.

\paragraph{Private protocol.}
We split each question's annotated gold evidence at random between the sharer
and the receiver, so that each model holds evidence the other cannot see, and we
report exact match. Every method compares against the same partition, the same
questions, and the same cached drafts.

\paragraph{Interventions.}
Matched, Deranged, and Receiver-only follow the definitions of
Appendix~\ref{app:existing-method-audit}, and
Appendix~\ref{app:our-intervention} gives the donor pools, permutations, and
seeds used for our own runs.

\paragraph{Aggregation.}
Every average over settings is unweighted: each model--benchmark cell counts
once regardless of its sample size, so the method means over the 35 Public
settings and the five-benchmark means in the scaling analysis are averages of
per-setting values, not sample-weighted pools. Counts such as ``30 of 35
settings'' are over the same 35 cells.

\subsection{Compute and Software}
\label{app:setup-compute}

All our experiments were conducted on 96GB GPUs whose peak BF16 tensor-core
rate is \TFLOPS{}~TFLOP/s and which reach $139.43$~TFLOP/s on BF16 matmul,
running models up to 8B parameters under bfloat16 precision. Runs use
PyTorch~$2.6.0$ with CUDA~$12.4$, Transformers~$4.52.4$, and driver
$535.161.08$. Sharer
drafts are generated and cached once per configuration before training, so the
same drafts serve the training stages, the reported evaluations, and the
interventions.

\paragraph{Cost baselines.}
Inference cost for Draft-KV comprises the sharer's prefill and draft decoding,
the additional forward pass that extracts the draft states
(Appendix~\ref{app:method-shapes}), the projections, and the receiver's prefill
and decoding.
Sharer-only reads its answer off the same draft, generated from the same prompt
under the same 512-token cap (Appendix~\ref{app:setup-eval}), so draft
decoding is common to it and to Draft-KV rather than an increment of ours.
Text-to-Text consumes the identical draft as text, which makes it the baseline
that isolates the medium: it prefills the draft through every receiver layer,
whereas Draft-KV reads the message positions at four communication layers
through the projections of Equation~\ref{eq:projection}. On the Public
benchmarks the receiver is scored from logits at one position, so it performs a
single prefill and no autoregressive decoding; the Private benchmarks add the
decoding of a short answer.

\paragraph{Measured cost per question.}
Table~\ref{tab:inference-cost} decomposes one question's inference for the
Qwen3-8B to Qwen2.5-0.5B-Instruct pair under both protocols. Drafts are
generated online rather than read from the cache, so draft decoding enters
every condition that consumes a draft.
Draft-KV performs $2.02\times$ the arithmetic of Sharer-only and takes
$1.01\times$ its latency; on HotpotQA the two ratios are $2.03\times$ and
$1.05\times$. The two ratios separate because the work the conditions share is
sequential and the work Draft-KV adds is not. Both decode the same draft,
$368$ memory-bound single-token steps on the 8B sharer that occupy $11.87$ of
Draft-KV's $12.02$~s. What Draft-KV adds is one compute-bound prefill over
prompt and draft: it doubles the sharer's arithmetic and costs $83$~ms. The
same decomposition separates
Draft-KV from Text-to-Text, which runs the identical sharer and leaves the
receiver $1.2$~ms earlier; the $6.56$ TFLOPs between the two conditions is the
extraction pass.

The interface is $0.88$ GFLOPs on MMLU-Redux and $0.32$ on HotpotQA, four
orders of magnitude below the total and $0.63$~ms of wall-clock. C2C occupies
the same position in the pipeline with $118.3$ GFLOPs and $35.6$~ms, so the
$348\times$ parameter ratio of Section~\ref{sec:introduction} reappears as a
$134\times$ ratio in the arithmetic each interface performs per question. C2C
also shows what the sharer's draft costs and buys: without draft decoding its
latency is $0.009\times$ Sharer-only, and Table~\ref{tab:main-results} places
its accuracy below the receiver's own.

Batch size one isolates per-question latency. At the batch sizes used for
evaluation, the same pair sustains $0.776$ questions per second under the
Public protocol at batch $16$ and $1.246$ under the Private protocol at
batch $8$.

\begin{table}[htbp]
\centering
\caption{Inference cost for one question with the Qwen3-8B sharer and the
Qwen2.5-0.5B-Instruct receiver. Stage latencies are medians over $200$
questions at batch size one after $20$ warmup questions, timed with CUDA
events. Floating-point counts are analytic; the interface column covers the
projections of Equation~\ref{eq:projection} and the external attention of
Equation~\ref{eq:app-external-attention}, and is reported in GFLOPs. Columns
marked $\times$ give the ratio to Sharer-only on the same benchmark. Dashes
mark stages a condition does not run.}
\label{tab:inference-cost}
\begingroup
\fontsize{7.6}{9.0}\selectfont
\setlength{\tabcolsep}{2.6pt}
\begin{tabular}{@{}lrrrrrrrrrrr@{}}
\toprule
& \multicolumn{6}{c}{Latency by stage (ms)}
& \multicolumn{2}{c}{Total latency}
& \multicolumn{3}{c}{Arithmetic} \\
\cmidrule(lr){2-7}\cmidrule(lr){8-9}\cmidrule(lr){10-12}
Condition
& \shortstack[r]{Sharer\\prefill}
& \shortstack[r]{Draft\\decode}
& \shortstack[r]{Extrac-\\tion}
& \shortstack[r]{Projec-\\tion}
& \shortstack[r]{Recv.\\prefill}
& \shortstack[r]{Recv.\\decode}
& ms & $\times$
& TFLOPs & $\times$ & \shortstack[r]{Interface\\GFLOPs} \\
\midrule
\multicolumn{12}{@{}l}{\textit{Public protocol, MMLU-Redux}} \\
Receiver-only  & --- & --- & --- & --- & 16.7 & --- & 16.7 & 0.001 & 0.108 & 0.016 & --- \\
Sharer-only    & 33.6 & 11{,}868.1 & --- & --- & --- & --- & 11{,}907.2 & 1.000 & 6.801 & 1.000 & --- \\
Text-to-Text   & 33.6 & 11{,}868.1 & --- & --- & 18.0 & --- & 11{,}925.2 & 1.002 & 7.187 & 1.057 & --- \\
Cache-to-Cache & 45.1 & --- & --- & 35.6 & 20.7 & --- & 101.4 & 0.009 & 2.256 & 0.332 & 118.3 \\
Draft-KV       & 33.6 & 11{,}868.1 & 83.3 & 0.6 & 19.2 & --- & 12{,}019.5 & 1.009 & 13.745 & 2.021 & 0.885 \\
\addlinespace
\multicolumn{12}{@{}l}{\textit{Private protocol, HotpotQA}} \\
Receiver-only  & --- & --- & --- & --- & 17.9 & 62.5 & 80.5 & 0.020 & 0.137 & 0.029 & --- \\
Sharer-only    & 46.4 & 3{,}935.7 & --- & --- & --- & --- & 3{,}981.7 & 1.000 & 4.687 & 1.000 & --- \\
Text-to-Text   & 46.4 & 3{,}935.7 & --- & --- & 18.4 & 61.9 & 4{,}150.8 & 1.042 & 4.924 & 1.050 & --- \\
Cache-to-Cache & \multicolumn{11}{c}{N/A} \\
Draft-KV       & 46.4 & 3{,}935.7 & 71.2 & 0.7 & 20.7 & 57.9 & 4{,}165.9 & 1.046 & 9.520 & 2.031 & 0.323 \\
\bottomrule
\end{tabular}
\endgroup
\end{table}

%% file: sections/appendix_ablation.tex
\subsection{Core Ablations: Results and Protocol}
\label{app:ablation-core}

\paragraph{Shared configuration.}
Every variant in this appendix is trained with the Qwen3-1.7B to
Qwen2.5-0.5B-Instruct configuration used for the main results: the same layer
assignment and interface capacity, the same per-stage data, optimizer settings,
and update budgets, the same four-to-one replay ratio in Stages~2 and~3, and
the same checkpoint-selection rule
(Appendix~\ref{app:experimental-setup}). Each variant is a single training run
and differs from Full Draft-KV only in the component its row names.

\paragraph{Evaluation and metrics.}
All results use the Public protocol and evaluate ARC-Challenge test
($n=1{,}172$) and MMLU-Redux test ($n=5{,}632$) with the option-scoring rule of
Section~\ref{sec:experimental-setup}. ARC test and MMLU-Redux are excluded from
Stage~3 training and checkpoint selection. The three message conditions and the
two gains are those of Section~\ref{sec:information-use-gap}, with Deranged drawing
its message from the same benchmark.
Table~\ref{tab:ablation-full} reports all three accuracies alongside $G$ and
$P$. One question is worth
$0.085$~pp on ARC-Challenge and $0.018$~pp on MMLU-Redux, which sets the
resolution of every difference discussed below.

\begin{table}[htbp]
\centering
\caption{Complete accuracy contrasts for the core ablations on the two
reported benchmarks. Accuracies are percentages; $G$ and $P$ are percentage
points. Differences are computed from the reported two-decimal accuracies.}
\label{tab:ablation-full}
\small
\setlength{\tabcolsep}{4pt}
\begin{tabular}{@{}lrrrrr@{}}
\toprule
Variant & $A_M$ & $A_D$ & $A_R$ & $G$ & $P$ \\
\midrule
\multicolumn{6}{@{}l}{\textit{ARC-Challenge test}} \\
Full Draft-KV & 83.53 & 37.63 & 40.10 & 43.43 & 45.90 \\
Prompt KV & 49.83 & 49.66 & 40.10 & 9.73 & 0.17 \\
w/o reconstruction pretraining & 76.71 & 37.97 & 40.10 & 36.61 & 38.74 \\
w/o answer alignment & 80.03 & 38.31 & 40.10 & 39.93 & 41.72 \\
\midrule
\multicolumn{6}{@{}l}{\textit{MMLU-Redux test}} \\
Full Draft-KV & 65.07 & 36.84 & 37.45 & 27.62 & 28.23 \\
Prompt KV & 43.59 & 43.50 & 37.45 & 6.14 & 0.09 \\
w/o reconstruction pretraining & 52.40 & 35.82 & 37.45 & 14.95 & 16.58 \\
w/o answer alignment & 55.40 & 36.15 & 37.45 & 17.95 & 19.25 \\
\bottomrule
\end{tabular}
\end{table}

\paragraph{Message-source control.}
Full Draft-KV exposes draft-position KV during task training and evaluation.
Prompt KV instead exposes only prompt-position KV in Stages~2 and~3 and at
evaluation. The receiver, communication architecture, Stage~1 reconstruction,
and subsequent reconstruction replay remain unchanged. Starting from the same
Stage~1 checkpoint, the Prompt KV interface is retrained for its task-message
source; this is not an inference-only replacement in the Full interface.
Replay continues to expose reconstruction-payload positions rather than prompt
positions. The two message sources also differ in how they are produced:
prompt-position states are available from the sharer's prefill, whereas
draft-position states require the decoding and extraction pass of
Appendix~\ref{app:method-shapes}. The comparison therefore tests message source
together with learning to use that source, not two encodings of equal length or
equal cost.

\paragraph{Stage-deletion controls.}
Without reconstruction pretraining, the interface enters Stage~2 without
Stage~1 updates and then proceeds to Stage~3, using the standard Xavier
initialization for projection matrices and zero initialization for gates
(Appendix~\ref{app:method-gates}).
Without answer alignment, the Stage~1 interface enters Stage~3 directly. Both
variants keep the replay schedule of Appendix~\ref{app:method-replay} in every
stage they retain: Stage~1, where it is kept, reconstructs on every update, and
Stages~2 and~3 alternate four task updates with one reconstruction update, so
removing pretraining does not remove reconstruction supervision altogether.

\paragraph{Effects of stage deletion.}
Beyond the Matched losses reported in Section~\ref{sec:ablation}, the reduced
pairing gains of both stage deletions do not translate into less mismatch harm:
$A_R-A_D$ decreases from $2.47$ to $2.13/1.79$~pp on ARC-C but increases from
$0.61$ to $1.63/1.30$~pp on MMLU-Redux.

\subsection{Guard Ablation: Mismatch Harm versus Pairing Gain}
\label{app:guard-ablation}

\paragraph{Control and diagnostic.}
The guard ablation branches from the Stage~2 checkpoint of Full Draft-KV and
sets $\lambda_p=0$ for Stage~3, leaving every other training condition
unchanged; Full uses $\lambda_p=0.1$ and $\tau=0.1$ as in
Equation~\ref{eq:app-stage3-loss}. Besides accuracy, we report the mean
per-example guard hinge on the test split under Deranged messages and the rate
at which it is active:
\begin{equation}
\bar h=\frac{1}{n}\sum_{i=1}^{n}
  \big[N_i^D-N_i^R-\tau\big]_+,
\qquad
\nu=\frac{100}{n}\sum_{i=1}^{n}
  \mathbf{1}\{N_i^D-N_i^R>\tau\}.
\label{eq:ablation-guard-diagnostics}
\end{equation}
Here $N_i^D$ and $N_i^R$ are full-vocabulary teacher-forced NLLs averaged over
supervised answer tokens, including the terminal token and excluding prompt
and padding positions (Equation~\ref{eq:app-token-nll}). We average the hinge
over examples, not over all tokens. $\bar h$ is measured in nats per target token;
$\nu$ is the percentage of examples exceeding the threshold. Stop-gradient on
$N_i^R$ affects training gradients but not these evaluation values.

\begin{table}[htbp]
\centering
\caption{Guard ablation. Accuracies are percentages; $G$, $P$, and
$A_R-A_D$ are in pp. $\bar h$ is the mean hinge and $\nu$ the threshold-exceedance
rate, with the number of affected examples in parentheses. Receiver-only
baselines are $40.10\%$ (ARC-C) and $37.45\%$ (MMLU-R).}
\label{tab:ablation-guard}
\small
\setlength{\tabcolsep}{3pt}
\begin{tabular}{@{}llrrrrrrr@{}}
\toprule
Variant & Benchmark & $A_M$ & $A_D$ & $G$ & $P$ & $A_R-A_D$ & $\bar h$ & $\nu$ \\
\midrule
Full & ARC-C & 83.53 & 37.63 & 43.43 & 45.90 & 2.47 & 0.038 & 19.7 (231) \\
w/o Guard & ARC-C & 82.94 & 35.92 & 42.84 & 47.02 & 4.18 & 0.091 & 37.9 (444) \\
\midrule
Full & MMLU-R & 65.07 & 36.84 & 27.62 & 28.23 & 0.61 & 0.052 & 24.6 (1386) \\
w/o Guard & MMLU-R & 64.68 & 36.24 & 27.23 & 28.44 & 1.21 & 0.104 & 42.8 (2411) \\
\bottomrule
\end{tabular}
\end{table}

\paragraph{Interpretation.}
Removing the guard lowers Matched accuracy by $0.59/0.39$~pp on
ARC-C/MMLU-Redux and increases mismatch harm by $1.71/0.60$~pp. The mean hinge
roughly doubles, from $0.038$ to $0.091$ and from $0.052$ to $0.104$, and the
exceedance rate does the same. The guard therefore limits mismatch harm while
retaining the Matched gains it was meant to preserve. The asymmetry between
benchmarks follows the training data: the guard is applied on Stage~3 ARC
updates, so its effect is strongest on ARC-C and carries over only partly to
MMLU-Redux.

This also shows why the pairing gain cannot be read on its own. Since
\begin{equation}
P=G+(A_R-A_D),
\label{eq:ablation-gain-decomposition}
\end{equation}
removing the guard raises $P$ to $47.02/28.44$~pp even as $G$ falls, because
the Deranged condition deteriorates faster than Matched improves. A larger
pairing gap obtained this way is damage under mismatch, not better
communication. The intervention tests cross-question mismatches and does not
speak to robustness against incorrect drafts for the same question.

\begin{figure}[htbp]
    \centering
    \includegraphics[width=\linewidth]{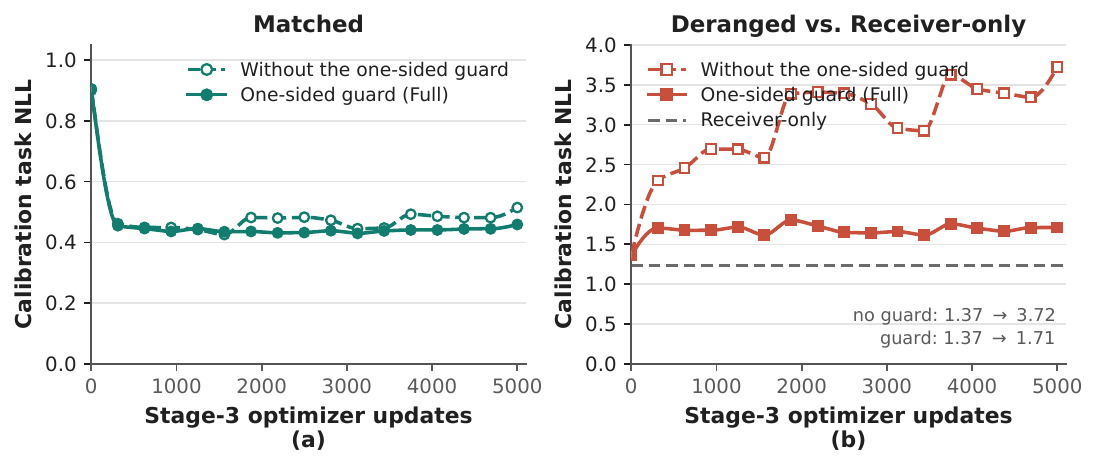}
    \caption{Stage-3 training dynamics with and without the one-sided guard,
    Qwen3-1.7B$\to$Qwen2.5-0.5B-Instruct, evaluated every 250 task updates on
    the 637-example calibration split. (a) Matched task NLL is comparable in
    the two runs. (b) Deranged task NLL against the constant Receiver-only NLL
    (dashed grey): without the guard Deranged NLL rises far past
    Receiver-only, while the one-sided guard keeps it close throughout
    training.}
    \label{fig:guard-training-curves}
\end{figure}

\paragraph{Training dynamics with and without the guard.}
Figure~\ref{fig:guard-training-curves} logs Stage-3 calibration NLL every 250
task updates in the run that keeps the one-sided guard ($\lambda_p=0.1$,
$\tau=0.1$, the Full row above) and in the run without it; the two share the
Qwen3-1.7B$\to$Qwen2.5-0.5B-Instruct pair, the Stage-1/Stage-2 initialization,
the ARC data, and the replay schedule, so their curves are directly comparable.
Matched NLL is essentially identical in the two runs (panel a), but Deranged
NLL diverges to $3.72$ nats without the guard---nearly $2.5$ nats above
Receiver-only---whereas the guard holds it at $1.71$, within $0.48$ nats of
Receiver-only (panel b). With no term bounding it, the gold-answer NLL under a
mismatched packet rises throughout training; the one-sided guard removes
exactly that drift while leaving Matched training untouched.

\begin{figure}[htbp]
    \centering
    \includegraphics[width=0.82\linewidth]{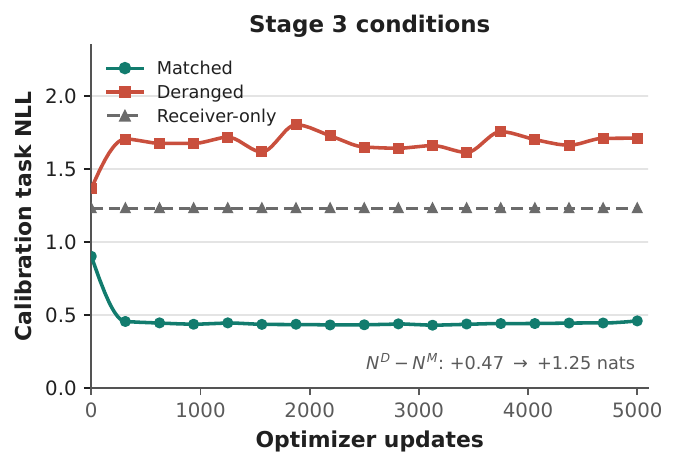}
    \caption{Stage-3 conditions of the reported model at the same calibration
    snapshots as Figure~\ref{fig:guard-training-curves}. Markers are
    measurements; curves are shape-preserving interpolation between them.}
    \label{fig:stage3-conditions}
\end{figure}

\paragraph{Pairing-gap dynamics.}
Figure~\ref{fig:stage3-conditions} follows the three Stage-3 conditions of Full
Draft-KV on shared axes: Matched task NLL falls from $0.90$ to $0.46$ nats
while Deranged rises from $1.37$ to $1.71$, so $N^D-N^M$ widens from
$0.47$ to $1.25$ nats over training. The separation is acquired rather than
optimized (Section~\ref{sec:stage3}): the guard bounds Deranged instead of
rewarding the gap, which is what keeps it near Receiver-only in
Figure~\ref{fig:guard-training-curves}.

%% file: sections/appendix_scaling.tex
Figure~\ref{fig:sharer-scaling} varies the sharer while holding everything else
fixed.

\paragraph{Fixed conditions.}
All four points share the frozen Qwen2.5-0.5B-Instruct receiver, the same layer
assignment, and an interface of identical size: as
Table~\ref{tab:setup-interface} shows, every Qwen3 sharer has a flattened KV
width of $1024$, so all four interfaces contain exactly $1{,}048{,}584$ trainable
parameters. They also share the Stage~3 training data, the update budget and
learning rates of Appendix~\ref{app:setup-training}, the seed, the
checkpoint-selection rule, the 512-token draft cap, and the five evaluation
benchmarks. One interface is trained per sharer; the curve therefore measures
system scaling at fixed interface capacity, not a single interface into which
sharers are swapped without training. The band in
Figure~\ref{fig:sharer-scaling} is the range across the five benchmarks, not a
confidence interval.

\paragraph{Capability and draft length.}
A larger sharer could help simply by writing more, since a longer draft means
more transmitted positions and more sharer computation at inference. The
measured lengths do not order the gains.
Table~\ref{tab:scaling-draft-length}
reports the mean number of decoded draft tokens per question, excluding the
sharer prompt and counting EOS where the draft terminates. Length is not
monotone in sharer size: Qwen3-4B writes \emph{shorter} drafts
than Qwen3-1.7B on all five benchmarks, $223$ tokens against $262$ on average,
yet its mean system gain is $8.57$~pp higher. Across the extremes the two
quantities are also out of proportion: from $0.6$B to $8$B
the average draft grows $5.2\times$ while the mean system gain grows from
$9.63$ to $42.25$~pp. With the conditions above held fixed, what distinguishes
the four points is the capability of the sharer writing the draft.

The $0.6$B sharer is the one case where length and capability are hard to
separate. Its drafts are short and highly variable, with a standard deviation
exceeding the mean on MMLU-Redux and C-EVAL, which reflects drafts that stop
early rather than a deliberately terse style. Its weak gains are consistent with
both a weaker draft and a shorter one.

\begin{table}[htbp]
\centering
\caption{Mean draft length in decoded tokens (mean $\pm$ SD) for the four
scaling points, all decoded greedily under a 512-token cap. The last column is
the unweighted mean over the five benchmarks, shown against the mean system
gain of Figure~\ref{fig:sharer-scaling}.}
\label{tab:scaling-draft-length}
\small
\setlength{\tabcolsep}{3.5pt}
\begin{tabular}{@{}lccccccr@{}}
\toprule
Sharer & MMLU-R & ARC-E & ARC-C & OBQA & C-EVAL & Mean & $\overline G$ (pp) \\
\midrule
Qwen3-0.6B & $85.3\pm88.1$ & $49.5\pm33.8$ & $58.5\pm41.9$ & $24.4\pm27.0$ & $73.2\pm93.0$ & 58.2 & 9.63 \\
Qwen3-1.7B & $293.1\pm112.9$ & $219.9\pm73.0$ & $243.4\pm81.7$ & $220.7\pm66.3$ & $332.1\pm120.0$ & 261.8 & 30.57 \\
Qwen3-4B & $263.5\pm108.4$ & $198.2\pm59.5$ & $212.9\pm69.4$ & $199.8\pm58.7$ & $242.2\pm128.0$ & 223.3 & 39.14 \\
Qwen3-8B & $358.9\pm102.3$ & $263.6\pm67.0$ & $285.3\pm76.7$ & $251.5\pm60.4$ & $355.8\pm113.3$ & 303.0 & 42.25 \\
\bottomrule
\end{tabular}
\end{table}

\paragraph{Other pairs.}
The remaining pairs in Table~\ref{tab:main-results} produce drafts in the same
range, averaging $233$ tokens for the Llama-3.2-3B-Instruct sharer and $303$ for
Qwen2.5-0.5B-Instruct in the reversed direction, so no configuration owes its
result to an unusually long message. The Qwen3-8B to Qwen3-4B pair reuses the
Qwen3-8B drafts.

%% file: sections/appendix_cases.tex
A receiver that relayed the sharer's answer would reproduce Sharer-only
accuracy exactly, so what separates communication from relay is how often the
receiver departs from the sharer, and in which direction. We count those
departures here; Appendix~\ref{app:case-examples}
then shows the individual questions behind those counts. All results use
Qwen3-1.7B as the sharer and Qwen2.5-0.5B-Instruct as the
receiver, on ARC-Challenge test and MMLU-Redux test under the Public protocol.
Every condition sees the same questions and the same cached drafts, so the
comparisons below are per-question rather than between aggregates.

\subsection{Disagreement with the Sharer}
\label{app:case-counts}

For each question we record whether a method is correct and whether the answer
extracted from the sharer's own draft is correct, which partitions the evaluation
set into four groups. Aggregating the partition recovers the columns of
Table~\ref{tab:main-results}; what it adds is the joint outcome, which those
columns cannot express. On ARC-Challenge, Draft-KV answers $979$ questions
correctly and Sharer-only $931$, a net difference of $48$ that is consistent
with repairing $48$ answers and breaking none, and equally with repairing
several hundred while breaking almost as many. Only the joint counts separate
these, and they also let us compare Draft-KV with Text-to-Text on the
identical draft, where the transmitted content is fixed and only the medium
differs.

\begin{figure}[htbp]
    \centering
    \includegraphics[width=\linewidth]{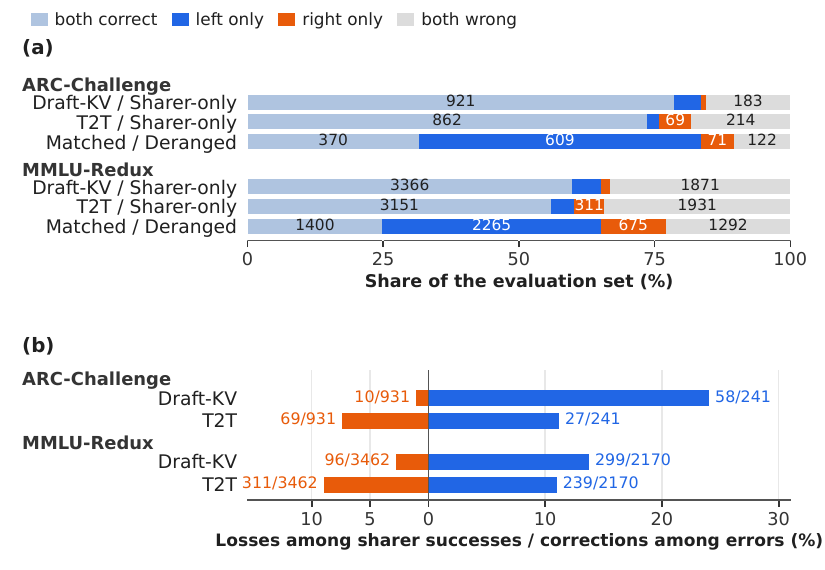}
    \caption{Per-question outcomes for every pairing, with counts shown
    throughout. (a) The complete partition as a share of the evaluation set;
    ``left'' and ``right'' refer to the two conditions named in each row.
    (b) The two disagreement cells of the method--sharer pairings, rescaled by
    the subsets they can occur in: corrections are only possible where the
    sharer erred, losses only where it did not. Draft-KV and T2T receive the
    same drafts.}
    \label{fig:case-outcomes}
\end{figure}

Figure~\ref{fig:case-outcomes}(a) gives the full partition for all three
pairings, and Figure~\ref{fig:case-outcomes}(b) rescales the two disagreement
cells by the subsets in which they can occur, since a correction is only
possible on a question the sharer got wrong and a loss only on one it got right.
On ARC-Challenge, Draft-KV recovers $58$ of the $241$ questions the sharer
answers incorrectly ($24.1\%$) while losing $10$ of the $931$ it answers
correctly ($1.1\%$); on MMLU-Redux the figures are $299$ of $2{,}170$ ($13.8\%$)
and $96$ of $3{,}462$ ($2.8\%$). Transcription would correct nothing, since it
reproduces Sharer-only by construction.
Recovering roughly a quarter of the sharer's
ARC-Challenge errors at a cost of about one percent of its successes is
available only to a model that reads the draft without deferring to it.

Comparing the two media is more informative than comparing either with the
sharer, because the draft is identical and only its representation changes.
Draft-KV and Text-to-Text correct a similar share of the sharer's MMLU-Redux
errors, $13.8\%$ against $11.0\%$, but they differ sharply in what they cost:
Text-to-Text loses $8.98\%$ of the sharer's correct answers against Draft-KV's
$2.77\%$, and on ARC-Challenge $7.41\%$ against $1.07\%$. Reading the draft as
text is thus roughly as good at supplying missing answers and three to seven
times more likely to overwrite correct ones, which is why Text-to-Text ends up
below Sharer-only overall ($-3.58$ and $-1.28$~pp) while Draft-KV ends up above
it ($+4.10$ and $+3.60$~pp). The advantage of the latent channel here is less
that it transmits more and more that the receiver is not talked out of answers it
already had right.

The same partition applied to the intervention makes the pairing gain concrete at
the level of individual questions, as the bottom row of each group in
Figure~\ref{fig:case-outcomes}(a) shows. Of the questions Draft-KV answers
correctly under Matched, $609$ of $979$ on ARC-Challenge ($62.2\%$) and $2{,}265$
of $3{,}665$ on MMLU-Redux ($61.8\%$) become incorrect once the message is
replaced by another question's. Content dependence is therefore not confined to
the aggregate: for most individual questions the correct answer requires the
message computed for that question.

Finally, the two benchmarks differ in the direction the training setup predicts.
Stage~3 trains on ARC, and it is on ARC-Challenge that the correction rate is
highest and the loss rate lowest; on MMLU-Redux, which no stage trains on,
corrections are less frequent and losses more than twice as likely. Using a draft
selectively is thus partly task-specific, though it transfers well enough to
leave a large net gain on the held-out benchmark.

Appendix~\ref{app:case-examples} examines representative questions from each
cell of this partition and from the split-evidence evaluation, tracing the
mechanisms behind the correction, medium, mismatch, and loss rates reported
above.

%% file: sections/appendix_baseline.tex
This appendix records how the audited methods were reproduced and which
implementation checks we ran.

\paragraph{C2C.}
We use the authors' released implementation and their published fuser for the
Qwen2.5-0.5B-Instruct to Qwen3-0.6B direction, trained on OpenHermes~2.5 under
the authors' released recipe with both backbones frozen and only the 28
per-layer projectors updated. We run it evaluation-only, without CoT, decoding
greedily with at most 64 new tokens and thinking disabled, and score
multiple-choice options at the assistant prefix
\texttt{The correct answer is}. Because no released fuser exists for the other
six directions of Table~\ref{tab:main-results}, we trained those ourselves under
the same recipe: one of the seven C2C rows rests on the authors' own weights and
the remaining six on our reproduction of their training. We also trained the
Qwen2.5-0.5B-Instruct$\to$Qwen3-0.6B direction ourselves under the same recipe;
those checkpoints supply the interval evidence of
Table~\ref{tab:c2c-bootstrap}, while Table~\ref{tab:main-results} and
Figure~\ref{fig:information-use-gap}(a) report the released fuser.

\paragraph{C2C reproduction check.}
To verify that our retraining follows the released recipe, we retrained C2C
fusers for the three sharer--receiver pairings evaluated in the original
paper, all with Qwen3-0.6B as the receiver. Table~\ref{tab:c2c-repro} compares
the reported accuracies with ours: the two agree within $0.44$~pp on every
benchmark, so the C2C entries in Table~\ref{tab:main-results} rest on a
pipeline that reproduces their training.

\begin{table}[htbp]
\centering
\caption{C2C accuracies reported in the original paper versus our retraining
under their released recipe, with Qwen3-0.6B as the receiver.}
\label{tab:c2c-repro}
\footnotesize
\setlength{\tabcolsep}{5pt}
\begin{tabular}{@{}llcc@{}}
\toprule
Sharer & Benchmark & Reported & Retrained \\
\midrule
\multirow{4}{*}{Qwen2.5-0.5B-Instruct} & MMLU-Redux & 42.92 & 42.95 \\
 & OpenBookQA & 52.60 & 52.80 \\
 & ARC-Challenge & 54.52 & 54.18 \\
 & C-EVAL & 41.77 & 41.68 \\
\addlinespace
\multirow{4}{*}{Llama-3.2-1B} & MMLU-Redux & 44.42 & 43.98 \\
 & OpenBookQA & 47.80 & 47.40 \\
 & ARC-Challenge & 53.39 & 53.41 \\
 & C-EVAL & 40.77 & 40.56 \\
\addlinespace
\multirow{4}{*}{Qwen3-4B-Base} & MMLU-Redux & 43.95 & 43.98 \\
 & OpenBookQA & 53.20 & 53.60 \\
 & ARC-Challenge & 55.39 & 55.03 \\
 & C-EVAL & 42.79 & 43.02 \\
\bottomrule
\end{tabular}
\end{table}

Within the protocol of Appendix~\ref{app:existing-method-audit}, the
substitution is applied after the sharer forward pass and after token and layer
alignment, but immediately before \texttt{C2CProjector} runs. The receiver
cache, the projector parameters, and the receiver's attention path are untouched,
so Matched and Deranged differ only in which question produced the source cache.

\paragraph{LatentMAS.}
LatentMAS is training-free, so we reproduce it from the authors' repository with
no adapter or projector to load. We evaluate the sequential
Planner--Critic--Refiner--Judger topology with a Qwen3-4B agent on a fixed
200-question GSM8K sample, where the latent message is the key--value cache an
agent accumulates over its $K$ latent steps and passes into the next agent's
forward pass. We report two runs: $K=20$ under greedy decoding with a 4096-token
cap, and $K=40$ under temperature $0.6$ and top-$p$ $0.95$ with a 2048-token
cap, each with its own seed. An earlier $K=20$ run at a 2048-token cap was
discarded because truncation affected its answers. The substitution is applied
after an agent produces that cache and before the next agent or the Judger
consumes it. Donor caches are length-matched to the target by truncating surplus
tail positions or zero-padding when shorter, without shifting the positions that
remain; the target prompt and decoding settings are unchanged.

\paragraph{Dense latent communication.}
No implementation of \emph{See What I See, Know What I
Think}~\citep{chen2026seewhatisee} has been released, so we do not reproduce
it. Every DLC
number we report, including the 8B and 14B points in
Figure~\ref{fig:information-use-gap}(c) and the parameter count in
Section~\ref{sec:introduction}, is taken from the published paper and is labelled
as such wherever it appears. We therefore make no claim about DLC's behaviour
under message intervention.

\paragraph{Implementation checks.}
Three checks were run on every configuration reported in this paper. First,
disabling communication reproduces the native receiver exactly: the maximum
absolute logit difference between the Receiver-only condition and the receiver
run alone is zero, so Receiver-only is the receiver's own function rather than a
near-miss approximation. Second, every donor assignment is verified to cover the
pool exactly once and to contain no fixed point, so no example can receive its
own message under Deranged. Third, for C2C we verify that the fused cache used
by the adapter-only control reconstructs the Matched cache to within $10^{-6}$
once its two compensation terms are restored, which confirms that the control
enters the model through the same path as Matched.

\paragraph{Uncertainty on the C2C pairing gains.}
Intervals are reported for four configurations of fusers we trained ourselves
under the authors' recipe---the
Qwen2.5-0.5B-Instruct$\to$Qwen3-0.6B fuser at training steps 2054 and 4109 and
at its final checkpoint, and a Llama-3.2-3B$\to$Qwen2.5-0.5B-Instruct fuser---for which
we computed per-question 95\% paired bootstrap intervals on the
Matched--Deranged difference over the five Public benchmarks (MMLU-Redux,
ARC-Easy, ARC-Challenge, OpenBookQA, C-Eval).
Table~\ref{tab:c2c-bootstrap} lists all twenty intervals. Thirteen contain
zero; the seven that do not are three positive intervals of the Llama fuser
(the largest point estimate, $+2.80$~pp, on OpenBookQA), two slightly negative
intervals of the step-2054 checkpoint, and one negative and one positive
interval of the final checkpoint, so even the intervals that
exclude zero remain within a few points of zero. The released C2C checkpoint
and the LatentMAS runs are covered by point estimates only.

\begin{table}[htbp]
\centering
\caption{Matched--Deranged differences (percentage points) with 95\%
per-question paired bootstrap intervals, for four configurations of C2C fusers
we trained ourselves under the authors' recipe. The first three columns are
training checkpoints (step 2054, step 4109, final) of a
Qwen2.5-0.5B-Instruct$\to$Qwen3-0.6B fuser; the fourth is a
Llama-3.2-3B-Instruct$\to$Qwen2.5-0.5B-Instruct fuser. The
Qwen2.5-0.5B-Instruct$\to$Qwen3-0.6B row of Table~\ref{tab:main-results} and
Figure~\ref{fig:information-use-gap}(a) instead report the authors' released
fuser. Intervals that exclude zero are bolded; thirteen of the twenty contain
zero.}
\label{tab:c2c-bootstrap}
\footnotesize
\setlength{\tabcolsep}{3.5pt}
\begin{tabular}{@{}llcccc@{}}
\toprule
Benchmark & & step 2054 & step 4109 & final & Llama$\to$0.5B\\
\midrule
\multirow{2}{*}{MMLU-R} & $\Delta$ & $\mathbf{-0.43}$ & $+0.11$ & $\mathbf{-0.47}$
 & $\mathbf{+1.07}$\\
 & CI & $\mathbf{[-.80,-.07]}$ & $[-.34,+.55]$ & $\mathbf{[-.91,-.03]}$
 & $\mathbf{[+.50,+1.65]}$\\
\addlinespace
\multirow{2}{*}{ARC-E} & $\Delta$ & $-0.31$ & $+0.35$ & $-0.09$
 & $\mathbf{+2.28}$\\
 & CI & $[-.88,+.26]$ & $[-.22,+.92]$ & $[-.70,+.52]$
 & $\mathbf{[+1.19,+3.38]}$\\
\addlinespace
\multirow{2}{*}{ARC-C} & $\Delta$ & $+0.26$ & $-0.35$ & $+0.42$ & $+1.13$\\
 & CI & $[-.61,+1.13]$ & $[-1.22,+.52]$ & $[-.53,+1.38]$ & $[-.17,+2.52]$\\
\addlinespace
\multirow{2}{*}{OBQA} & $\Delta$ & $+1.00$ & $-0.40$ & $+0.00$
 & $\mathbf{+2.80}$\\
 & CI & $[-.20,+2.40]$ & $[-1.80,+1.00]$ & $[-1.10,+1.10]$
 & $\mathbf{[+.60,+5.20]}$\\
\addlinespace
\multirow{2}{*}{C-Eval} & $\Delta$ & $\mathbf{-0.32}$ & $+0.09$ & $\mathbf{+0.74}$
 & $-0.19$\\
 & CI & $\mathbf{[-.58,-.06]}$ & $[-.22,+.40]$ & $\mathbf{[+.45,+1.03]}$
 & $[-.61,+.21]$\\
\bottomrule
\end{tabular}
\end{table}

%% file: sections/appendix_case_study.tex
\definecolor{CaseRule}{HTML}{C9CED4}    
\definecolor{CaseGreen}{HTML}{2E7D32}   
\definecolor{CaseGreenBack}{HTML}{F3F9F3}
\definecolor{CaseBlue}{HTML}{1F5FA8}    
\definecolor{CaseBlueBack}{HTML}{F2F6FB}
\definecolor{CaseViolet}{HTML}{6B3FA0}  
\definecolor{CaseVioletBack}{HTML}{F7F4FB}
\definecolor{CaseRed}{HTML}{B3261E}     
\definecolor{CaseRedBack}{HTML}{FBF4F3}
\definecolor{CaseTeal}{HTML}{00695C}    
\definecolor{CaseTealBack}{HTML}{F2F8F6}

\newcommand{\cgood}{\textcolor{CaseGreen}{\ensuremath{\checkmark}}}
\newcommand{\cbad}{\textcolor{CaseRed}{\ensuremath{\times}}}

\newtcolorbox{casebox}[3]{%
  breakable, enhanced,
  colback=#3, colframe=#2,
  boxrule=0.5pt, arc=1.6pt,
  left=5pt, right=5pt, top=6pt, bottom=5pt,
  toptitle=1pt, bottomtitle=1pt,
  fonttitle=\sffamily\bfseries\fontsize{7.6}{8.8}\selectfont,
  coltitle=black, colbacktitle=#3,
  title={#1},
  attach boxed title to top left={xshift=6pt, yshift=-\tcboxedtitleheight/2},
  boxed title style={colback=white, colframe=#2, boxrule=0.5pt, arc=1pt,
                     left=3pt, right=3pt, top=0.7pt, bottom=0.7pt}}

\newtcolorbox{draftbox}{%
  enhanced, colback=white, colframe=CaseRule,
  boxrule=0.35pt, arc=1pt,
  left=4pt, right=3.5pt, top=3pt, bottom=3pt}

This appendix steps from the aggregate partition of
Appendix~\ref{app:case-counts} down to the individual questions behind it.
All cases use the setting of Appendix~\ref{app:case-counts}: Qwen3-1.7B as
the sharer and Qwen2.5-0.5B-Instruct as the receiver (Stage-3 checkpoint),
on ARC-Challenge and MMLU-Redux test under the Public protocol, plus four
examples (Cases~16--19) from the split-evidence evaluation on HotpotQA and
2WikiMultihopQA. Within each group,
cases are selected by a fixed rule (ascending sample identifier). Drafts are
excerpted with elisions marked; the receiver always sees the full draft.

The four multiple-choice groups instantiate the cells of the partition:
\emph{corrected} (sharer wrong, Draft-KV right), \emph{medium} (same draft,
T2T wrong, Draft-KV right), \emph{mismatch} (matched message right, deranged
message wrong), and \emph{lost} (sharer right, Draft-KV wrong). The
split-evidence cases are the generation counterpart: each model sees only
half of the evidence, so the two halves must be composed through the message.

\subsection{Corrected Cases}
\label{app:case-corrected}

Corrected cases are questions the sharer answers wrongly and Draft-KV answers
correctly. The seven examples below split into three mechanisms. In Cases
1--3 the receiver already knows the answer: the latent message leaves it
intact while the text medium talks it out of it. In Cases 4--5 the
receiver's prior is strong enough that no condition moves it, which is the
limiting case of the same protective behaviour. In Cases 6--7 every other
condition fails---receiver alone, sharer, and the same draft read as
text---so the answer enters only through the paired message: correction as
instruction rather than preservation.

\begin{casebox}{Case 1 \;$\cdot$\; corrected \;$\cdot$\; ARC-Challenge}{CaseGreen}{CaseGreenBack}
\small\setlength{\parskip}{3.2pt}
\textbf{Question.} A student pours salt water into a beaker and then pours
some acid into another beaker. The student returns to the lab table and
realizes that the samples are not marked. What should the student do?

\textbf{Choices.} (A)~smell each sample in order to identify it;
(B)~guess which beaker contains each substance;
(C)~mix the substances together to determine what they are;
(D)~safely discard the samples and get new ones.

\textbf{Gold.} (D)~safely discard the samples and get new ones.

\textbf{Sharer draft (excerpt).}
\begin{draftbox}
\footnotesize\itshape
When a student pours salt water into a beaker and then pours some acid into
another beaker, and the samples are not marked, the best approach is to avoid
mixing the substances together, as this could lead to unintended reactions or
contamination. Option A (smell each sample) is not reliable, as many
substances are odorless or have similar smells.~[\dots]~The correct answer is
LETTER C.
\end{draftbox}

\begin{center}
\begin{tabular}{@{}lccccc@{}}
\toprule
 & Sharer & Recv-only & T2T & Draft-KV & Deranged\\
\midrule
 & C~\cbad & D~\cgood\,{\scriptsize(.559)} & C~\cbad &
 D~\cgood\,{\scriptsize(.440)} & A~\cbad\,{\scriptsize(.361)}\\
\bottomrule
\end{tabular}
\end{center}

\textbf{Analysis.} The draft reasons \emph{against} mixing the substances,
which is the option it then selects, and closes on the wrong letter; T2T
reproduces that letter, while Draft-KV keeps the receiver's own correct
answer. The deranged message, by contrast, collapses the decision and pulls
the receiver to a third option.

\end{casebox}

\begin{casebox}{Case 2 \;$\cdot$\; corrected \;$\cdot$\; ARC-Challenge}{CaseGreen}{CaseGreenBack}
\small\setlength{\parskip}{3.2pt}
\textbf{Question.} Wolves, which are top predators, were eliminated from
Yellowstone National Park in the 1930s. In 1995, wolves were reintroduced
into Yellowstone. During the period in which wolves were absent from
Yellowstone, which most likely occurred?

\textbf{Choices.} (A)~an increase in competition for food resources among
small prey; (B)~a greater opportunity for primary producers to flourish;
(C)~an increase in the population of tertiary consumers;
(D)~a greater balance of predator-prey relationships.

\textbf{Gold.} (A)~an increase in competition for food resources among small
prey.

\textbf{Sharer draft (excerpt).}
\begin{draftbox}
\footnotesize\itshape
When wolves are removed from an ecosystem, they play a key role as top
predators. They help control the populations of their prey, which in turn
affects the availability of resources for other species. Without wolves, prey
populations may increase, leading to overgrazing and a decrease in the growth
of primary producers (like plants).~[\dots]~The correct answer is LETTER B.
\end{draftbox}

\begin{center}
\begin{tabular}{@{}lccccc@{}}
\toprule
 & Sharer & Recv-only & T2T & Draft-KV & Deranged\\
\midrule
 & B~\cbad & A~\cgood\,{\scriptsize(.643)} & B~\cbad &
 A~\cgood\,{\scriptsize(.367)} & D~\cbad\,{\scriptsize(.981)}\\
\bottomrule
\end{tabular}
\end{center}

\textbf{Analysis.} The draft's own opening reasoning---prey increase, then
overgrazing---points away from its closing letter, and T2T copies the closing
letter anyway; Draft-KV keeps the receiver's correct answer at reduced
confidence. The deranged message is worse than no message at all: the
receiver commits to a wrong option with probability $.981$.

\end{casebox}

\begin{casebox}{Case 3 \;$\cdot$\; corrected \;$\cdot$\; ARC-Challenge}{CaseGreen}{CaseGreenBack}
\small\setlength{\parskip}{3.2pt}
\textbf{Question.} A student walks to school one morning and notices the
grass is wet but the streets are dry. Which of these processes most likely
caused the grass to be wet?

\textbf{Choices.} (A)~condensation; (B)~erosion; (C)~evaporation;
(D)~precipitation.

\textbf{Gold.} (A)~condensation.

\textbf{Sharer draft (excerpt).}
\begin{draftbox}
\footnotesize\itshape
To determine which process most likely caused the grass to be wet, let's
analyze the situation step by step: the grass is wet, and the streets are
dry.~[\dots]~The most likely process that caused the grass to be wet is
precipitation, as it is the process that brings water from the sky to the
grass, making it wet. The correct answer is LETTER D.
\end{draftbox}

\begin{center}
\begin{tabular}{@{}lccccc@{}}
\toprule
 & Sharer & Recv-only & T2T & Draft-KV & Deranged\\
\midrule
 & D~\cbad & A~\cgood\,{\scriptsize(.988)} & D~\cbad &
 A~\cgood\,{\scriptsize(.508)} & B~\cbad\,{\scriptsize(.722)}\\
\bottomrule
\end{tabular}
\end{center}

\textbf{Analysis.} The draft's conclusion cannot explain the observation
(rain wets the streets as well), and T2T adopts it verbatim. Draft-KV retains
the receiver's near-certain answer, though the matched probability falls from
$.988$ to $.508$: the message transmits the sharer's pull without the
sharer's conclusion winning.

\end{casebox}

\begin{casebox}{Case 4 \;$\cdot$\; corrected \;$\cdot$\; MMLU-Redux (anatomy)}{CaseGreen}{CaseGreenBack}
\small\setlength{\parskip}{3.2pt}
\textbf{Question.} A patient has a fracture in which the radius is bent but
not displaced, and the skin is intact. This type of fracture is known as
which of the following?

\textbf{Choices.} (A)~Closed, greenstick; (B)~Complex, comminuted;
(C)~Compound, transverse; (D)~Open, spiral.

\textbf{Gold.} (A)~Closed, greenstick.

\textbf{Sharer draft (excerpt).}
\begin{draftbox}
\footnotesize\itshape
To determine the correct answer, let's break down the key features of the
fracture described: the radius is bent but not displaced; the skin is intact,
which means the fracture is closed, not open (compound).~[\dots]~The correct
answer is LETTER C.
\end{draftbox}

\begin{center}
\begin{tabular}{@{}lccccc@{}}
\toprule
 & Sharer & Recv-only & T2T & Draft-KV & Deranged\\
\midrule
 & C~\cbad & A~\cgood\,{\scriptsize(.968)} & A~\cgood &
 A~\cgood\,{\scriptsize(.545)} & A~\cgood\,{\scriptsize(.671)}\\
\bottomrule
\end{tabular}
\end{center}

\textbf{Analysis.} The draft states the decisive fact---skin intact means
closed---and then closes on an option whose first word, compound, means open.
Every receiver condition stays with (A). The message reshapes confidence
($.968 \to .545$) but not the decision: when the receiver knows, the channel
does not argue.

\end{casebox}

\begin{casebox}{Case 5 \;$\cdot$\; corrected \;$\cdot$\; MMLU-Redux (marketing)}{CaseGreen}{CaseGreenBack}
\small\setlength{\parskip}{3.2pt}
\textbf{Question.} Which type of retailer involves comparatively low prices
as a major selling point combined with the reduced costs of doing business?

\textbf{Choices.} (A)~Discount retailers.; (B)~Convenience stores.;
(C)~Category killer stores.; (D)~Limited line retailers.

\textbf{Gold.} (A)~Discount retailers.

\textbf{Sharer draft (excerpt).}
\begin{draftbox}
\footnotesize\itshape
To answer this question, let's break down the key elements: ``low prices as a
major selling point'' indicates a focus on affordability; ``reduced costs of
doing business'' suggests that the retailer operates with lower operational
costs.~[\dots]~The best fit is C. Category killer stores, as they combine low
prices with reduced business costs through their focus on a specific
category. The correct answer is LETTER C.
\end{draftbox}

\begin{center}
\begin{tabular}{@{}lccccc@{}}
\toprule
 & Sharer & Recv-only & T2T & Draft-KV & Deranged\\
\midrule
 & C~\cbad & A~\cgood\,{\scriptsize(.997)} & A~\cgood &
 A~\cgood\,{\scriptsize(.507)} & A~\cgood\,{\scriptsize(.669)}\\
\bottomrule
\end{tabular}
\end{center}

\textbf{Analysis.} As in Case 4, the receiver answers correctly in every
condition, and the matched message costs about half of the probability mass
($.997 \to .507$) without changing the decision. The interface attends to the
message, but a wrong conclusion inside it does not transfer.

\end{casebox}

\begin{casebox}{Case 6 \;$\cdot$\; corrected \;$\cdot$\; MMLU-Redux (abstract algebra)}{CaseGreen}{CaseGreenBack}
\small\setlength{\parskip}{3.2pt}
\textbf{Question.} Statement 1: $4x - 2$ is irreducible over $\mathbb{Z}$.
Statement 2: $4x - 2$ is irreducible over $\mathbb{Q}$.

\textbf{Choices.} (A)~True, True; (B)~False, False; (C)~True, False;
(D)~False, True.

\textbf{Gold.} (D)~False, True.

\textbf{Sharer draft (excerpt).}
\begin{draftbox}
\footnotesize\itshape
To determine the truth of the statements, we need to analyze the
irreducibility of the polynomial $4x - 2$ over $\mathbb{Z}$ and
$\mathbb{Q}$. A polynomial is irreducible over $\mathbb{Z}$ if it cannot be
factored into the product of two non-constant polynomials with integer
coefficients. Let's try to factor: $4x - 2 = 2(2x - 1)$.~[\dots]~Since $2$
is a constant polynomial (degree 0), it is not considered a non-constant
polynomial. Therefore, $4x - 2$ is reducible over $\mathbb{Z}$.~[\dots]~The
correct answer is LETTER B.
\end{draftbox}

\begin{center}
\begin{tabular}{@{}lccccc@{}}
\toprule
 & Sharer & Recv-only & T2T & Draft-KV & Deranged\\
\midrule
 & B~\cbad & A~\cbad\,{\scriptsize(.488)} & B~\cbad &
 D~\cgood\,{\scriptsize(.505)} & B~\cbad\,{\scriptsize(.923)}\\
\bottomrule
\end{tabular}
\end{center}

\textbf{Analysis.} Both statements turn on the same factorization, $4x - 2
= 2(2x - 1)$: over $\mathbb{Z}$ the factor 2 is not invertible, so this is a
genuine factorization (False); over $\mathbb{Q}$ it is invertible, so it is
not (True). The draft writes the factorization down but never invokes
invertibility, concludes ``reducible'' twice, and lands on (B); the receiver
alone answers (A); T2T follows the draft. Under the matched message the
receiver picks (D)---the pairing that the draft's own factorization supports
and its verdict contradicts. Content, again, travels; conclusions do not.

\end{casebox}

\begin{casebox}{Case 7 \;$\cdot$\; corrected \;$\cdot$\; MMLU-Redux (high school US history)}{CaseGreen}{CaseGreenBack}
\small\setlength{\parskip}{3.2pt}
\textbf{Question.} ``Permit us, then, earnestly to recommend these articles
to the immediate and dispassionate attention of the legislatures of the
respective states.~[\dots]~under a conviction of the absolute necessity of
uniting all our councils and all our strength, to maintain and defend our
common liberties\ldots'' (Journals of the Continental Congress, 1777). The
most notable achievement of the United States under the Articles of
Confederation was\ldots

\textbf{Choices.} (A)~the creation of a strong executive office to lead the
national government; (B)~the empowerment of Congress to regulate commerce;
(C)~the empowerment of Congress to collect taxes; (D)~the provision for
land sales in the Northwest that would benefit the entire nation.

\textbf{Gold.} (D)~the provision for land sales in the Northwest that would
benefit the entire nation.

\textbf{Sharer draft (excerpt).}
\begin{draftbox}
\footnotesize\itshape
The Articles created a weak central government with no strong executive, so
this is incorrect. The Articles did not give Congress the power to collect
taxes, which was a major weakness of the system. The Articles did not
provide for land sales in the Northwest, and such provisions were not part
of the original document.~[\dots]~The correct answer is LETTER C.
\end{draftbox}

\begin{center}
\begin{tabular}{@{}lccccc@{}}
\toprule
 & Sharer & Recv-only & T2T & Draft-KV & Deranged\\
\midrule
 & C~\cbad & A~\cbad\,{\scriptsize(.719)} & A~\cbad &
 D~\cgood\,{\scriptsize(.350)} & A~\cbad\,{\scriptsize(.503)}\\
\bottomrule
\end{tabular}
\end{center}

\textbf{Analysis.} The draft correctly rules out (A)---no strong executive
under the Articles---and (C)---no power to tax---then denies the one
provision the Articles did make, the Northwest land ordinances, and, having
rejected (A), (C), and the gold (D), closes on the very (C) it had just
excluded. The
receiver alone answers (A), and T2T follows. The matched message selects
(D), at the lowest confidence in this group ($.350$): the receiver lands on
the single option that the sharer's draft dismissed without grounds.

\end{casebox}

The seven cases span the two ends of correction: in Cases 1--5 the receiver
already holds the answer and the latent message declines to overrule it, where
text transmitted the sharer's error three times out of three; in Cases 6--7 no
condition but the paired message produces the answer at all.

\subsection{Medium Cases}
\label{app:case-medium}

Medium cases hold the draft fixed and vary only its representation: the same
question, the same sharer draft, T2T wrong and Draft-KV right. They isolate
what the latent medium does with content that the text medium demonstrably
fails to deliver.

\begin{casebox}{Case 8 \;$\cdot$\; medium \;$\cdot$\; ARC-Challenge}{CaseBlue}{CaseBlueBack}
\small\setlength{\parskip}{3.2pt}
\textbf{Question.} How does the speed of radio waves compare with the speed
of infrared waves?

\textbf{Choices.} (A)~Radio waves travel faster.; (B)~Infrared waves travel
faster.; (C)~They both travel at the same speed.; (D)~The wave speed varies
with their frequency.

\textbf{Gold.} (C)~They both travel at the same speed.

\textbf{Sharer draft (excerpt).}
\begin{draftbox}
\footnotesize\itshape
To answer this question, we need to understand the nature of radio waves and
infrared waves. Radio waves are a type of electromagnetic wave, and all
electromagnetic waves travel at the speed of light in a
vacuum.~[\dots]~Therefore, both radio waves and infrared waves travel at the
same speed in a vacuum. The correct answer is LETTER C.
\end{draftbox}

\begin{center}
\begin{tabular}{@{}lccccc@{}}
\toprule
 & Sharer & Recv-only & T2T & Draft-KV & Deranged\\
\midrule
 & C~\cgood & A~\cbad\,{\scriptsize(.758)} & A~\cbad &
 C~\cgood\,{\scriptsize(.925)} & B~\cbad\,{\scriptsize(.926)}\\
\bottomrule
\end{tabular}
\end{center}

\textbf{Analysis.} The receiver's prior is confidently wrong ($.758$ on
``radio waves travel faster''). Reading the same draft as text leaves the
error untouched; reading it as key--value states reverses it into $.925$
confidence in the right answer. The deranged message is equally decisive in
the wrong direction, which underlines that the matched gain is carried by
content, not by the mere presence of the channel.

\end{casebox}

\begin{casebox}{Case 9 \;$\cdot$\; medium \;$\cdot$\; MMLU-Redux (high school mathematics)}{CaseBlue}{CaseBlueBack}
\small\setlength{\parskip}{3.2pt}
\textbf{Question.} How many diagonals are in a regular octagon?

\textbf{Choices.} (A)~8; (B)~16; (C)~17; (D)~20.

\textbf{Gold.} (D)~20.

\textbf{Sharer draft (excerpt).}
\begin{draftbox}
\footnotesize\itshape
To find the number of diagonals in a regular octagon, we can use the formula
for the number of diagonals in an $n$-sided polygon, $n(n-3)/2$. For an
octagon, $n = 8$: $8 \times 5 / 2 = 40/2 = 20$. The correct answer is LETTER
D.
\end{draftbox}

\begin{center}
\begin{tabular}{@{}lccccc@{}}
\toprule
 & Sharer & Recv-only & T2T & Draft-KV & Deranged\\
\midrule
 & D~\cgood & A~\cbad\,{\scriptsize(.729)} & A~\cbad &
 D~\cgood\,{\scriptsize(.992)} & B~\cbad\,{\scriptsize(.705)}\\
\bottomrule
\end{tabular}
\end{center}

\textbf{Analysis.} The draft contains a complete derivation, yet its text
version does not move the receiver off $8$---the number of vertices, a
natural confusion for a small model. The latent version moves it to $.992$ on
the right value. Whatever the message encodes, it behaves more like the
result of the sharer's computation than like its prose.

\end{casebox}

\begin{casebox}{Case 10 \;$\cdot$\; medium \;$\cdot$\; MMLU-Redux (medical genetics)}{CaseBlue}{CaseBlueBack}
\small\setlength{\parskip}{3.2pt}
\textbf{Question.} The \_\_\_\_\_\_ is the set of observable characteristics
and is the sum of genetic and environmental effects.

\textbf{Choices.} (A)~genotype; (B)~phenotype; (C)~both genotype and
phenotype; (D)~neither genotype or phenotype.

\textbf{Gold.} (B)~phenotype.

\textbf{Sharer draft (excerpt).}
\begin{draftbox}
\footnotesize\itshape
To answer this question, let's break it down: genotype refers to the genetic
makeup of an organism, including all the genes it carries. Phenotype refers
to the observable characteristics of an organism, which result from the
interaction between the genotype and the environment.~[\dots]~The correct
answer is LETTER B.
\end{draftbox}

\begin{center}
\begin{tabular}{@{}lccccc@{}}
\toprule
 & Sharer & Recv-only & T2T & Draft-KV & Deranged\\
\midrule
 & B~\cgood & B~\cgood\,{\scriptsize(.817)} & A~\cbad &
 B~\cgood\,{\scriptsize(.982)} & A~\cbad\,{\scriptsize(.879)}\\
\bottomrule
\end{tabular}
\end{center}

\textbf{Analysis.} The receiver starts out correct at $.817$. The text medium
talks it out of the right answer---the draft discusses the genotype first,
and T2T lands there---while the latent medium raises the same receiver to
$.982$. This is the loss asymmetry of Appendix~\ref{app:case-counts} in
miniature: text drafts overwrite correct receivers, latent drafts do not.

\end{casebox}

\subsection{Mismatch Cases}
\label{app:case-mismatch}

Mismatch cases keep the interface and the question fixed and swap only the
message: the matched message yields the right answer, the deranged message
the wrong one. They are the per-question form of the content-dependence
result in Appendix~\ref{app:case-counts}.

\begin{casebox}{Case 11 \;$\cdot$\; mismatch \;$\cdot$\; ARC-Challenge}{CaseViolet}{CaseVioletBack}
\small\setlength{\parskip}{3.2pt}
\textbf{Question.} An atom of beryllium has 4 protons, 4 electrons, and 5
neutrons. What is its mass number?

\textbf{Choices.} (A)~4; (B)~5; (C)~8; (D)~9.

\textbf{Gold.} (D)~9.

\textbf{Sharer draft (excerpt).}
\begin{draftbox}
\footnotesize\itshape
To determine the mass number of an atom, we need to add the number of protons
and neutrons in the nucleus: mass number $= 4 + 5 = 9$. The correct answer is
LETTER D.
\end{draftbox}

\begin{center}
\begin{tabular}{@{}lccccc@{}}
\toprule
 & Sharer & Recv-only & T2T & Draft-KV & Deranged\\
\midrule
 & D~\cgood & A~\cbad\,{\scriptsize(.677)} & D~\cgood &
 D~\cgood\,{\scriptsize(.996)} & A~\cbad\,{\scriptsize(.999)}\\
\bottomrule
\end{tabular}
\end{center}

\textbf{Deranged donor.} ``There are four seasons in a year. Which reason is
most responsible for the changing seasons on Earth?'' (gold A: the way Earth
tilts on its axis).

\textbf{Analysis.} Both receiver readings are nearly certain---$.996$ on the
right answer with the matched message, $.999$ on the wrong one with the
donor's---and the only thing that changed is the message. Note also that the
deranged prediction, (A), is the donor's gold letter.

\end{casebox}

\begin{casebox}{Case 12 \;$\cdot$\; mismatch \;$\cdot$\; MMLU-Redux (elementary mathematics)}{CaseViolet}{CaseVioletBack}
\small\setlength{\parskip}{3.2pt}
\textbf{Question.} Solve the equation $14 = w + 23$.

\textbf{Choices.} (A)~$-37$; (B)~$-9$; (C)~9; (D)~37.

\textbf{Gold.} (B)~$-9$.

\textbf{Sharer draft (excerpt).}
\begin{draftbox}
\footnotesize\itshape
To solve the equation $14 = w + 23$, we need to isolate the variable $w$.
Subtract 23 from both sides: $14 - 23 = w$, so $w = -9$. The correct answer is
LETTER B.
\end{draftbox}

\begin{center}
\begin{tabular}{@{}lccccc@{}}
\toprule
 & Sharer & Recv-only & T2T & Draft-KV & Deranged\\
\midrule
 & B~\cgood & A~\cbad\,{\scriptsize(.649)} & B~\cgood &
 B~\cgood\,{\scriptsize(.686)} & C~\cbad\,{\scriptsize(.928)}\\
\bottomrule
\end{tabular}
\end{center}

\textbf{Deranged donor.} ``In the movie `Star Wars' what kind of creature is
Chewbacca?'' (gold C: Wookiee).

\textbf{Analysis.} The matched message turns a wrong $.649$ into a correct
$.686$; the donor's message turns the same receiver into $.928$ confidence on
(C)---again the donor's gold letter, and here the sign-flipped version of the
right value.

\end{casebox}

\begin{casebox}{Case 13 \;$\cdot$\; mismatch \;$\cdot$\; MMLU-Redux (public relations)}{CaseViolet}{CaseVioletBack}
\small\setlength{\parskip}{3.2pt}
\textbf{Question.} In what year did the BBC start broadcasting radio?

\textbf{Choices.} (A)~1917; (B)~1922; (C)~1925; (D)~1927.

\textbf{Gold.} (B)~1922.

\textbf{Sharer draft (excerpt).}
\begin{draftbox}
\footnotesize\itshape
The BBC started broadcasting radio in 1922. This is a well-documented fact,
as the BBC's first radio broadcast was on 4 February 1922, marking the
beginning of its radio broadcasting services. The correct answer is LETTER B.
\end{draftbox}

\begin{center}
\begin{tabular}{@{}lccccc@{}}
\toprule
 & Sharer & Recv-only & T2T & Draft-KV & Deranged\\
\midrule
 & B~\cgood & A~\cbad\,{\scriptsize(.526)} & B~\cgood &
 B~\cgood\,{\scriptsize(.910)} & D~\cbad\,{\scriptsize(.512)}\\
\bottomrule
\end{tabular}
\end{center}

\textbf{Deranged donor.} ``A 2008 survey showed that what percentage of the
world's largest companies are reporting their corporate responsibility?''
(gold D: 80\%).

\textbf{Analysis.} The receiver's prior is barely above chance ($.526$); the
matched message resolves it to $.910$ on 1922. The deranged message resolves
it to (D)---the donor's gold letter, for the third time in three mismatch
cases.

\end{casebox}

In all three cases the deranged prediction coincides with the gold letter of
the donor question. Three cases cannot separate a message that carries
answer identity from one that carries content, and the ascending-identifier
selection rule was not designed to test that distinction.
Appendix~\ref{app:adapter-only} settles it at scale: averaging the interface
output over five donors drops accuracy below Receiver-only, which a channel
transporting letters alone would not do.

\subsection{Lost Cases}
\label{app:case-lost}

Lost cases are the cost side of the partition: the sharer is right and
Draft-KV is wrong. We show one from each benchmark.

\begin{casebox}{Case 14 \;$\cdot$\; lost \;$\cdot$\; ARC-Challenge}{CaseRed}{CaseRedBack}
\small\setlength{\parskip}{3.2pt}
\textbf{Question.} The morning temperature in a city is $41^{\circ}$F. If a
sunny, mild day is forecast, which temperature is most likely for 2:00 p.m.?

\textbf{Choices.} (A)~$32^{\circ}$F; (B)~$41^{\circ}$F; (C)~$78^{\circ}$F;
(D)~$98^{\circ}$F.

\textbf{Gold.} (C)~$78^{\circ}$F.

\textbf{Sharer draft (excerpt).}
\begin{draftbox}
\footnotesize\itshape
To determine the most likely temperature at 2:00 p.m., we need to consider
typical temperature changes on a sunny, mild day: a typical increase can be
around 10--20$^{\circ}$F, so the temperature is most likely in the range of
51--61$^{\circ}$F. Among the choices, 78$^{\circ}$F is significantly higher
than this range~[\dots]~The most reasonable choice is C. 78$^{\circ}$F. The
correct answer is LETTER C.
\end{draftbox}

\begin{center}
\begin{tabular}{@{}lccccc@{}}
\toprule
 & Sharer & Recv-only & T2T & Draft-KV & Deranged\\
\midrule
 & C~\cgood & B~\cbad\,{\scriptsize(.702)} & C~\cgood &
 B~\cbad\,{\scriptsize(.583)} & B~\cbad\,{\scriptsize(.653)}\\
\bottomrule
\end{tabular}
\end{center}

\textbf{Analysis.} The draft contradicts itself---it computes a
51--61$^{\circ}$F range, calls $78^{\circ}$F significantly higher, and then
selects $78^{\circ}$F anyway---and the message evidently transmits the
hesitation rather than the conclusion: the receiver stays on its own prior
(B, the unchanged morning temperature), with the matched message shaving its
confidence from $.702$ to $.583$ but not flipping it. Here the full text
draft does what the message cannot, and T2T recovers the right answer.

\end{casebox}

\begin{casebox}{Case 15 \;$\cdot$\; lost \;$\cdot$\; MMLU-Redux (high school mathematics)}{CaseRed}{CaseRedBack}
\small\setlength{\parskip}{3.2pt}
\textbf{Question.} Suppose $f(x)$ is an invertible function, and suppose
that $f(2)=f^{-1}(2)=4$. What is the value of $f(f(2))$?

\textbf{Choices.} (A)~1; (B)~2; (C)~4; (D)~0.

\textbf{Gold.} (B)~2.

\textbf{Sharer draft (excerpt).}
\begin{draftbox}
\footnotesize\itshape
We are given that $f(x)$ is an invertible function with $f(2) = f^{-1}(2) =
4$. First, $f(2) = 4$ means that when $x = 2$, $f(x) = 4$; and $f^{-1}(2) =
4$ implies $f(4) = 2$.~[\dots]~We know $f(2) = 4$, so $f(f(2)) = f(4) = 2$.
The correct answer is B.
\end{draftbox}

\begin{center}
\begin{tabular}{@{}lccccc@{}}
\toprule
 & Sharer & Recv-only & T2T & Draft-KV & Deranged\\
\midrule
 & B~\cgood & A~\cbad\,{\scriptsize(.473)} & B~\cgood &
 A~\cbad\,{\scriptsize(.290)} & C~\cbad\,{\scriptsize(.463)}\\
\bottomrule
\end{tabular}
\end{center}

\textbf{Analysis.} The answer requires chaining two facts the receiver cannot
check for itself. The receiver keeps its prior (A) under the matched message:
probability falls from $.473$ to $.290$, the lowest confidence in this
appendix, but the prediction does not change---as if the receiver sensed the
message pointing somewhere it cannot reconstruct. Text-to-Text succeeds
because the draft's derivation can be followed step by step; the message, by
contrast, is best at carrying answers, not the derivations behind them.

\end{casebox}

Both losses share a shape: the decisive content is a process---a
self-correcting estimate in Case 14, a two-step derivation in Case 15---that
the receiver has no way to verify against its own knowledge, and a
fixed-size message does not carry process as well as a full text draft does.
These are the roughly $1\%$ of ARC-Challenge and $3\%$ of MMLU-Redux sharer
successes that Draft-KV gives back in Appendix~\ref{app:case-counts}.

\subsection{Split-Evidence Generation Cases}
\label{app:case-split}

The last four cases come from the split-evidence evaluation: the question is
shown to both models, but the evidence is divided into two halves, one shown
to the sharer and one to the receiver, and neither half alone suffices to
answer. The sharer is asked to reason over its half and close with an answer
line; the receiver keeps its half and is instructed to reply with a single
short answer, so the entry listed for each receiver condition is its complete
output; Text-to-Text is part of this protocol and its scores on the two
benchmarks are reported in Table~\ref{tab:main-results}. Answers are scored by
exact match, so a correct fact delivered as a full sentence counts as wrong.
In all four examples the sharer answers wrongly and Draft-KV answers correctly.

\begin{casebox}{Case 16 \;$\cdot$\; split-evidence \;$\cdot$\; HotpotQA}{CaseTeal}{CaseTealBack}
\small\setlength{\parskip}{3.2pt}
\textbf{Question.} The rapper whose debut album was titled ``Thug Misses''
has sold over how many records worldwide?

\textbf{Gold.} 2 million.

\textbf{Sharer evidence.} ``Thug Misses is the debut album by American
rapper Khia.''

\textbf{Receiver evidence.} ``To date Khia has collectively sold over 2
million records worldwide.''

\textbf{Sharer draft (full).}
\begin{draftbox}
\footnotesize\itshape
The rapper whose debut album was titled ``Thug Misses'' is Khia. According
to sales data, Khia's debut album ``Thug Misses'' has sold over 20 million
records worldwide.

Answer: 20 million
\end{draftbox}

\begin{center}
\begin{tabular}{@{}llc@{}}
\toprule
Sharer-only & 20 million & \cbad\\
Receiver-only & 200{,}000 & \cbad\\
Draft-KV (matched) & 2 million & \cgood\\
Deranged & Rapper Khia has sold over 2 million records worldwide. & \cbad\\
\bottomrule
\end{tabular}
\end{center}

\textbf{Analysis.} The halves are complementary: the sharer holds the link
(Thug Misses is Khia's debut album), the receiver holds the count (Khia has
sold 2 million records). Alone, the sharer inflates the count tenfold---its
half contains no sales figure at all---and the receiver, holding the right
figure but unable to attach it to the question, guesses $200{,}000$. Paired,
the answer is exact. The deranged message produces a third failure mode: the
receiver abandons the short-answer format and emits a full sentence that
merely restates its own evidence.

\end{casebox}

\begin{casebox}{Case 17 \;$\cdot$\; split-evidence \;$\cdot$\; HotpotQA}{CaseTeal}{CaseTealBack}
\small\setlength{\parskip}{3.2pt}
\textbf{Question.} The ``Happy Fun Ball'' was the subject of a series of
parody advertisements on a show created by who?

\textbf{Gold.} Lorne Michaels.

\textbf{Sharer evidence.} ``The `Happy Fun Ball' was the subject of a series
of parody advertisements on `Saturday Night Live'.''

\textbf{Receiver evidence.} ``Saturday Night Live (abbreviated as SNL) is an
American late-night live television sketch comedy and variety show created
by Lorne Michaels and developed by Dick Ebersol.''

\textbf{Sharer draft (full).}
\begin{draftbox}
\footnotesize\itshape
The ``Happy Fun Ball'' was the subject of a series of parody advertisements
on ``Saturday Night Live.''

Answer: Lorne Balfe
\end{draftbox}

\begin{center}
\begin{tabular}{@{}llc@{}}
\toprule
Sharer-only & Lorne Balfe & \cbad\\
Receiver-only & Saturday Night Live & \cbad\\
Draft-KV (matched) & Lorne Michaels & \cgood\\
Deranged & 1908 & \cbad\\
\bottomrule
\end{tabular}
\end{center}

\textbf{Analysis.} The answer sits in the receiver's half; the link to it
(Happy Fun Ball aired on Saturday Night Live) sits in the sharer's. The
sharer resolves the link correctly and then hallucinates the wrong creator
(Lorne Balfe is a film composer); the receiver, holding the true creator,
answers with the show's name because nothing tells it that its show is the
one being asked about. The matched message transmits the link rather than
the sharer's wrong conclusion, and the receiver finishes the chain with its
own half: the system is right where both halves, separately, are wrong.

\end{casebox}

\begin{casebox}{Case 18 \;$\cdot$\; split-evidence \;$\cdot$\; 2WikiMultihopQA}{CaseTeal}{CaseTealBack}
\small\setlength{\parskip}{3.2pt}
\textbf{Question.} Where was the husband of Lou Henry Hoover born?

\textbf{Gold.} West Branch, Iowa.

\textbf{Sharer evidence.} ``Lou Henry Hoover (born Lou Henry; March 29, 1874
-- January 7, 1944) was the wife of President of the United States Herbert
Hoover and served as the First Lady of the United States from 1929 to
1933.''

\textbf{Receiver evidence.} ``Born to a Quaker family in West Branch, Iowa,
he grew up in Oregon.''

\textbf{Sharer draft (full).}
\begin{draftbox}
\footnotesize\itshape
The husband of Lou Henry Hoover was President Herbert Hoover. Herbert Hoover
was born in West Virginia.

Answer: West Virginia
\end{draftbox}

\begin{center}
\begin{tabular}{@{}llc@{}}
\toprule
Sharer-only & West Virginia & \cbad\\
Receiver-only & Oregon & \cbad\\
Draft-KV (matched) & West Branch, Iowa & \cgood\\
Deranged & Oregon & \cbad\\
\bottomrule
\end{tabular}
\end{center}

\textbf{Analysis.} The receiver's half states the birthplace without naming
its subject; the sharer's half names the subject without the birthplace.
Alone, each model guesses: the sharer invents West Virginia, the receiver
picks the place of upbringing out of its own sentence. Under the deranged
message the output coincides with receiver-only---an incongruous message is
discarded, and the receiver falls back on its half. Only the matched message
composes the two halves.

\end{casebox}

\begin{casebox}{Case 19 \;$\cdot$\; split-evidence \;$\cdot$\; 2WikiMultihopQA}{CaseTeal}{CaseTealBack}
\small\setlength{\parskip}{3.2pt}
\textbf{Question.} Where was the place of death of the director of film Foma
Gordeyev (Film)?

\textbf{Gold.} Moscow.

\textbf{Sharer evidence.} ``Foma Gordeyev is a 1959 Soviet drama film
directed by Mark Donskoy.''

\textbf{Receiver evidence.} ``He worked in the script department, then as an
assistant director in Moscow, later as an editing assistant in Leningrad.''

\textbf{Sharer draft (full).}
\begin{draftbox}
\footnotesize\itshape
The director of the film Foma Gordeyev (1959) is Mark Donskoy. Based on the
provided evidence, the place of death of Mark Donskoy is not directly
stated. However, Mark Donskoy was a Soviet filmmaker who died in 1964.
Therefore, the place of death is likely in the Soviet Union, specifically in
a city where he was based during his career.

Answer: Soviet Union
\end{draftbox}

\begin{center}
\begin{tabular}{@{}llc@{}}
\toprule
Sharer-only & Soviet Union & \cbad\\
Receiver-only & Leningrad & \cbad\\
Draft-KV (matched) & Moscow & \cgood\\
Deranged & United States & \cbad\\
\bottomrule
\end{tabular}
\end{center}

\textbf{Analysis.} Neither half states where Donskoy died; the answer must
come from the models' parameters, with the halves supplying the chain (film
$\to$ director $\to$ career cities). The sharer follows the chain, concedes
that the deathplace is absent from its evidence, and falls back on the
country rather than a city. The
receiver's half names two cities and it picks the wrong one; the deranged
message yields a fourth answer; the matched message settles on the right one.

\end{casebox}